\documentclass[10pt,twocolumn,letterpaper]{article}

\usepackage{iccv}              

\definecolor{iccvblue}{rgb}{0.21,0.49,0.74}
\usepackage[pagebackref,breaklinks,colorlinks,allcolors=iccvblue]{hyperref}
\usepackage[accsupp]{axessibility}  

\usepackage{multirow}
\usepackage{graphicx}
\usepackage{amsmath}
\usepackage[normalem]{ulem}

\def\paperID{3670} 
\def\confName{ICCV}
\def\confYear{2025}

\title{BlinkTrack: Feature Tracking over 80 FPS via Events and Images}

\author{%
Yichen Shen\textsuperscript{1}\thanks{Yichen Shen, Yijin Li contributed equally to this work.}\quad
Yijin Li\textsuperscript{1,2}\footnotemark[1]\quad
Shuo Chen\textsuperscript{1}\quad
Guanglin Li\textsuperscript{1}\\
Zhaoyang Huang\textsuperscript{2}\quad
Hujun Bao\textsuperscript{1}\quad
Zhaopeng Cui\textsuperscript{1}\quad
Guofeng Zhang\textsuperscript{1}\thanks{Corresponding author.}\\[0.3em]
\textsuperscript{1}State Key Lab of CAD\&CG, Zhejiang University \qquad
\textsuperscript{2}Avolution AI\\
{\tt\small \{shenyichen,zhangguofeng\}@zju.edu.cn}
}

\begin{document}
\maketitle

\begin{abstract}
Event cameras, known for their high temporal resolution and ability to capture asynchronous changes, have gained significant attention for their potential in feature tracking, especially in challenging conditions. However, event cameras lack the fine-grained texture information that conventional cameras provide, leading to error accumulation in tracking. To address this, we propose a novel framework, BlinkTrack, which integrates event data with grayscale images for high-frequency feature tracking. Our method extends the traditional Kalman filter into a learning-based framework, utilizing differentiable Kalman filters in both event and image branches. 
This approach improves single-modality tracking and effectively solves the data association and fusion from asynchronous event and image data.
We also introduce new synthetic and augmented datasets to better evaluate our model. 
Experimental results indicate that BlinkTrack significantly outperforms existing methods, exceeding 80 FPS with multi-modality data and 100 FPS with preprocessed event data.
Codes and dataset are available at \url{https://github.com/ColieShen/BlinkTrack}.
\end{abstract}


\section{Introduction}
\label{sec:intro}

Feature tracking aims to estimate the trajectories of query points from a reference timestamp over subsequent periods. It serves as the cornerstone for many computer vision tasks, including structure from motion\cite{schoenberger2016sfm,schoenberger2016mvs}, simultaneous localization and mapping(SLAM)\cite{Mur_Artal_2015, murORB2,ORBSLAM3_TRO,tof_slam,pats}, object tracking\cite{Wojke2017simple,Wojke2018deep,zhang2021fairmot}, and multi-view stereo\cite{dong2024globaldepthrangefreemultiviewstereo, yao2018mvsnetdepthinferenceunstructured}.

In recent years, event cameras\cite{Gallego_2018,eklt,blinkflow,blinkvision,slide_gcn} have attracted significant attention within the research community, particularly in the field of feature tracking. Event cameras are innovative sensors that asynchronously detect changes in a scene with very high temporal resolution, capturing events as they occur rather than recording frames at fixed intervals. This capability allows event cameras to track features at high frequencies, even in challenging lighting conditions or with fast-moving objects, where conventional cameras are easy to fail\cite{Zhang_2023_CVPR, zhang2021object}.

\begin{figure}[tb]
  \centering
  \includegraphics[width=1\linewidth]{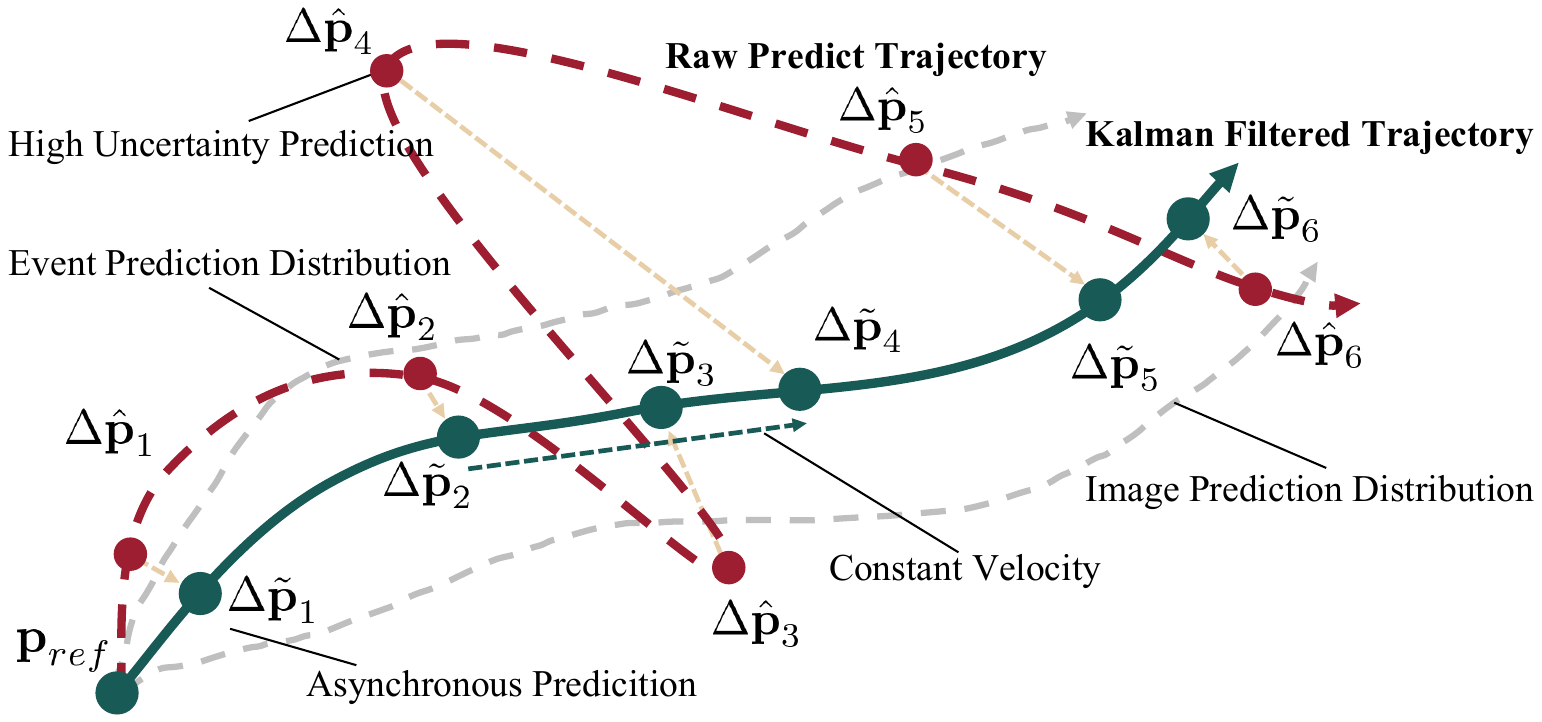}
  \caption{
  Our method employs the idea of the Kalman filter to associate the asynchronous data from event and image modality and learn the uncertainty-aware fusion.
  }
  \label{fig:teaser}
\end{figure}

Nevertheless, the event cameras have inherent limitations.
For example, event cameras are unable to capture detailed, fine-grained texture information like conventional cameras, and the spatial signal of event cameras could be very sparse in slow-moving scenes.
As a result, event-only could lead to error accumulation and inhibit tracking performance. 
Therefore, in this paper, we aim to explore efficient and powerful feature tracking by fusing the event data and the image frame from conventional cameras.

To achieve our objective, we must address three primary challenges:
(1) Data association. Event cameras and standard cameras do not operate in a synchronized manner, with standard cameras capturing at fixed frequency(e.g., 30 Hz\footnote{In this paper, we use FPS as the frequency of algorithm inference, and Hz as the frame rate of the input data.}) and event cameras detecting changes asynchronously. As a result, it needs an effective method to associate these asynchronous data streams from different modalities.
(2) Uncertainty-aware fusion. Event cameras and standard cameras exhibit different distributions of uncertainty due to their distinct sensor characteristics\cite{eventservey}. To achieve effective fusion, the method should be capable of adaptively weighting the contributions of each modality based on confidence estimates. In addition, the uncertainty caused by occlusion and other environmental factors also needs to be considered.
(3) Runtime efficiency. A promising event-based framework requires minimal runtime overhead to fully exploit the low latency and high temporal resolution from event camera data in scenarios demanding rapid response times, such as autonomous driving and drones.

Previous learning-based methods for feature tracking either adopt a naive fusion strategy\cite{deepev}, resulting in marginal performance gains, or align event data with low-frame-rate image data before fusion, thereby sacrificing the high temporal resolution inherent to event data.
In other fields\cite{Zhang_2023_CVPR, zhang2021object,deltar}, attention-based\cite{vaswani2023attention} modules have been explored for implicit alignment and fusion. While these methods can be effective, they do not meet the efficiency requirements for high-frame-rate feature tracking.
On the other hand, traditional techniques such as the Kalman filter\cite{10.1115/1.3662552} offer efficient tools for the fusion of asynchronous data. 
However, they typically require manually tuned parameters, such as carefully adjusting the uncertainty of observations, yet still underperform compared to recent learning-based methods.

Based on these observations, we propose a framework leveraging the Kalman filter technique and extending it into a learning-based paradigm.
In our framework, an event module and an image module are employed to independently generate initial tracking predictions from the input data. Additionally, we estimate the prediction uncertainty for each module. Then the Kalman filter integrates these network predictions(treated as observations) according to the respective uncertainties.
Since the Kalman filter maintains an internal motion state that evolves dynamically and can be updated with observations at any timestamp, it naturally supports associating the asynchronous predictions from two modules.
We train these networks coupled with the Kalman filter in an end-to-end fashion. Supervised by ground-truth tracking labels and proxy visibility labels, the networks learn to generate initial predictions and estimate uncertainties,
minimizing the loss by dynamically weighting from historical and current observations of different modalities and obtaining the reliable fused state with the Kalman filter (Fig.~\ref{fig:teaser}).

A remaining challenge is computational efficiency: while the Kalman filter is lightweight, the event and image module constitutes the primary computational bottleneck when computing the initial prediction.
To this end, we carefully modulate the event and image modules based on state-of-the-art methods\cite{dpvo,deepev,pips}, ensuring robust tracking while maintaining high computational efficiency.

We also find that existing datasets and benchmarks are overly simple. As a result, they do not fully enable our model to reach its potential or provide a comprehensive evaluation. To address these, we generate a more complex synthetic dataset for training. Additionally, we augment two existing evaluation datasets to introduce more challenging scenarios and ensure a more thorough evaluation.

Our contributions can be outlined as follows. 
First, we propose a Kalman-filter-based framework for feature tracking that effectively solves the data association and fusion from asynchronous event data and image data.
Second, we introduce the carefully modulated event and image module that strikes a good balance between efficiency and performance.
Third, we generate new datasets for training and evaluating event-based feature tracking.
Exclusive experiments show that the proposed methods outperform existing methods by a large margin and are much more robust in handling occlusions. Besides, it can run at 80 FPS on multi-modality data and over 100 FPS on preprocessed event data.

\section{Related Work}

\noindent{\textbf{Event-based Feature Tracking.}}
Most early works\cite{8491018,Dardelet2021AnEF,9093607,chui2021eventbased,ecdt,eventblob} are based on hand-crafted features or optimization.
Among them, HASTE\cite{haste} and EKLT\cite{eklt} are more well-known and open source.
Deep-EV-Tracker\cite{deepev} is the first to apply neural networks to event-based feature tracking.
It is trained on MultiFlow\cite{multiflow} after deriving the tracking labels from the ground truth optical flow, which leads to inaccurate tracking labels due to quantization error and occlusion.
A concurrent work Deep-Ev-Tracker(E2VID)\cite{deepev2} enhances Deep-Ev-Tracker by training with E2VID\cite{e2vid, e2vid2} data but requires significant preprocessing time.
Another concurrent work, ETAP\cite{han2024eventbasedtrackingpointmotionaugmented}, achieves excellent tracking performance by applying a CoTracker-like\cite{cotracker} network. However, its heavy model results in a long runtime, diminishing the benefits of high temporal resolution and low latency of event data.
Our event module is inspired by Deep-EV-Tracker\cite{deepev} but modulates the network architecture to achieve a balance between accuracy and efficiency.
Besides, we propose a new training dataset that provides accurate tracking ground truth for both occluded and non-occluded pixels, along with visibility labels.

\noindent{\textbf{Frame-based Feature Tracking.}}
PIPs\cite{pips} is the first learning-based method that directly predicts the tracking trajectory given an RGB sequence.
PIPs has inspired a lot of follow-up work\cite{contextpips,tapvid,pointodyssey}, among which the CoTracker\cite{cotracker} series is currently the most robust and widely used.
However, these methods are time- and resource-consuming, which are incompatible with the low-latency nature of event data. To address this, we design a lightweight image module that can seamlessly cooperate with our event module.

\begin{figure*}[tb]
  \centering
  \includegraphics[width=0.9\linewidth]{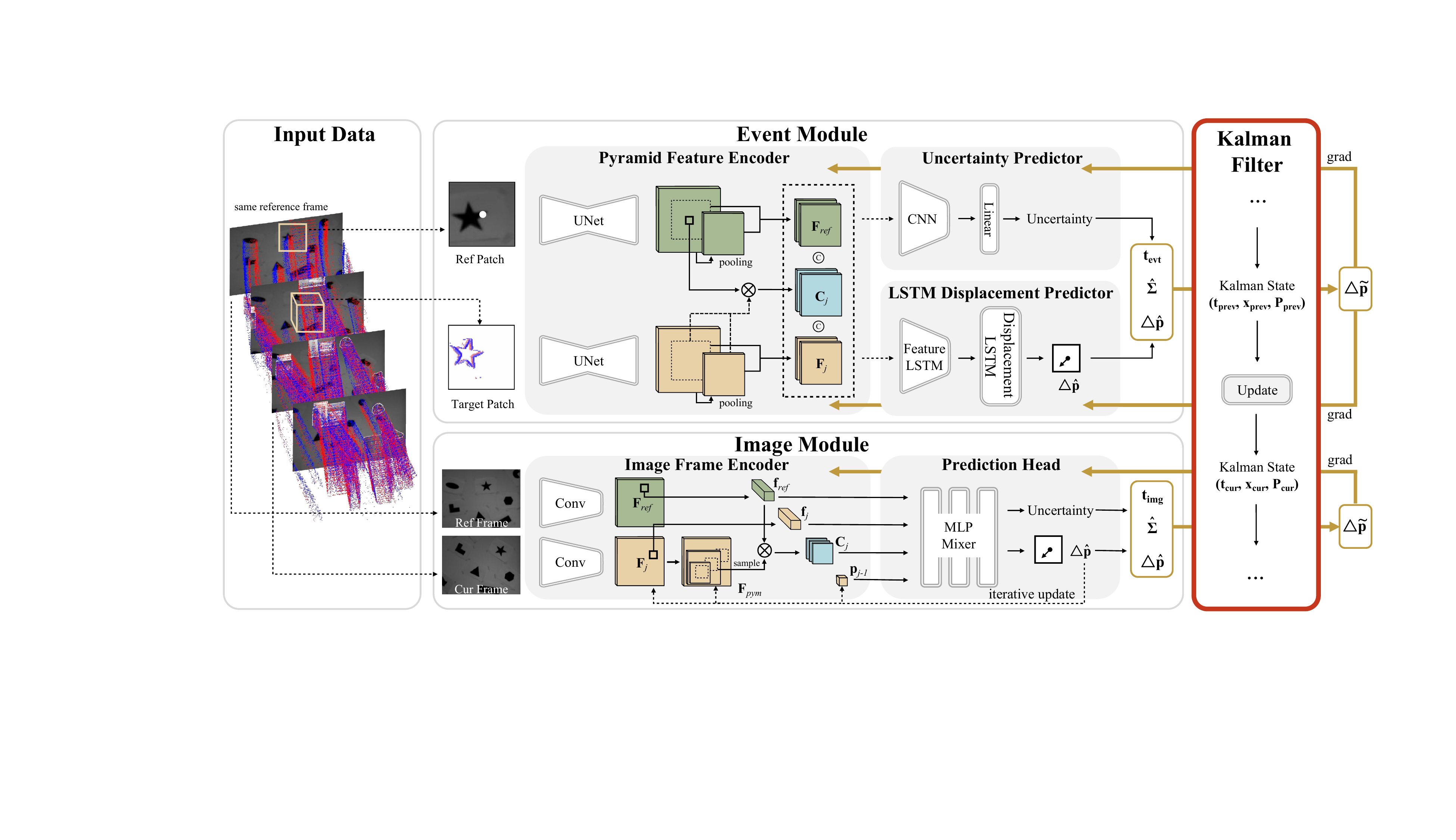}
  \caption{
      \textbf{Overview of the proposed framework.}
      BlinkTrack employs a Kalman filter to integrate an event module and an image module. The event module takes a reference patch $\mathbf{P}_{evt_{ref}}$ extracted from a grayscale image $\mathbf{I}_{ref}$ and an event patch $\mathbf{P}_{evt_j}$ preprocessed from the event stream at timestamp $\mathbf{t}_{evt_j}$ as input. Separate pyramid encoders first process the two patches to generate pyramid feature maps, which are then used to compute pyramid correlation maps. These feature and correlation maps are concatenated and fed to prediction heads(an uncertainty predictor and an LSTM displacement predictor) to obtain the uncertainty $\hat{\mathbf{\Sigma}}_{evt_j}$ and relative displacement $\Delta\hat{\mathbf{p}}$. The image module operates similarly but encodes the full frame and utilizes an MLP to aggregate diverse data for prediction. 
      Both module predictions are fed into the Kalman filter, which generates the final results $\Delta\tilde{\mathbf{p}}$ with supervised optimal uncertainty.
      The final integration achieves both high accuracy and efficiency, supporting asynchronous data fusion.
  }
  \label{fig:struct}
\end{figure*}

\noindent{\textbf{Event and Frame Fusion.}}
The fusion of event and image data has proven effective across various fields\cite{Zhang_2023_CVPR, zhang2021object, 9359329, 10.1007/978-3-031-72907-2_10, Pellerito2023DeepVO}. 
In the area of depth estimation, Gehrig et al.\cite{9359329} asynchronously encode event or image data into a shared latent feature space; the encoded features are then fused implicitly using an RNN-like predictor. However, the results are not satisfactory, which may be due to the trade-off between the fusion network's capacity and computational efficiency.
In the area of feature tracking,
Deep-EV-Tracker\cite{deepev} naively combines event data and images by initializing the tracking position with results from another modality, leading to limited improvements on its pure event tracking. 
A concurrent work, FF-KDT\cite{ffkdt}, accumulates event data to align it with the frame rate of image frames before applying a neural network for fusion. This approach avoids the need for data association but sacrifices the high temporal resolution of event data and results in a low frame rate output.
Instead, our method is inspired by the idea of the Kalman filter that effectively and efficiently solves the data association and fusion problem.

\noindent{\textbf{Kalman Filter.}}
Kalman filter has been widely used for object tracking\cite{Blackman1999DesignAA,10.1155/2012/870890, zhang2022bytetrack,yang2024samurai} and optical flow\cite{336247,566996} because it can build an explicit motion model, benefiting the tracking task. The differentiability of the Kalman filter makes it compatible with neural networks. Bao et al.\cite{8451564,8662710} add a Kalman filter on existing optical flow methods, gaining performance promotion. Also, the Kalman filter holds the advantage of fusing different signals, which is used by Wang et al.\cite{Wang_2021} to reconstruct video using LDR images and event data.
This paper investigates Kalman filter techniques in feature tracking and couples them with two modulated modules to develop a powerful and efficient framework.

\section{BlinkTrack}

Given a query point $\mathbf{p}_{ref}$ at a particular timestamp, BlinkTrack attempts to track the points in subsequent timestamps by matching them with the event stream and the image data. The initial tracks are predicted using the event module and image module, and then the asynchronous predictions from both modules are associated and fused using the Kalman filter technique with timestamp and supervised uncertainty, predicting final position $\tilde{\mathbf{p}} = \mathbf{p}_{ref} + \Delta\tilde{\mathbf{p}}$.
An overview of the proposed method, BlinkTrack, is presented in Fig.~\ref{fig:struct}.

\subsection{Event Module} \label{Event Tracking Module}
Our event module follows previous methods\cite{eklt,deepev,deepev2} to take a grayscale image patch as the reference and predict the track position by matching with the event patch.
The use of image patches as references ensures reference consistency with the image module.

\noindent{\textbf{Pyramid Feature Encoder.}}
To minimize computation, we extract the reference patch  $\mathbf{P}_{evt_{ref}}$ from the reference grayscale image $\mathbf{I}_{ref}$ at query position $\mathbf{p}_{ref}$, and the event patch $\mathbf{P}_{evt_{j}}$ from the preprocessed event frame at last predicted position $\mathbf{p}_{j-1}$.
Details of the preprocessing steps are provided in the supplementary materials. 
The reference and event patches are then encoded using two shallow U-Nets, which results in two patch feature maps.
To enhance the feature, inspired by DPVO\cite{dpvo}, we extract feature maps at different scales from the center of the full feature maps to construct a two-level feature pyramid, enabling multi-scale feature perception.
The feature pyramid is then reshaped to the same scale by applying average pooling and concatenated, generating the reference feature map $\mathbf{F}_{evt_{ref}}$ and the event patch feature map $\mathbf{F}_{evt_{j}}$.
Additionally, we compute the correlation $\mathbf{C}_{evt_j}$ correlated between the reference feature vector $\mathbf{f}_{evt_{ref}}$ and the event patch feature map $\mathbf{F}_{evt_{j}}$, explicitly comparing the similarity between the reference and event features. This reference feature vector $\mathbf{f}_{evt_{ref}}$ is extracted from the center of the reference feature map $\mathbf{F}_{evt_{ref}}$ as shown in Fig.~\ref{fig:struct}.

\noindent{\textbf{LSTM Displacement Predictor.}}
Previous methods\cite{gehrig2023recurrent,deepev} demonstrated that LSTMs significantly enhance event-based tasks by efficiently extracting temporal features and producing smoother results. Building on this, we employ dual LSTM modules for feature and displacement information propagation.
In the feature LSTM module, the $\mathbf{F}_{evt_{ref}}$, $\mathbf{F}_{evt_j}$, and $\mathbf{C}_{evt_j}$ are concatenated and convolved to enhance information exchange before being processed through the ConvLSTM block\cite{convlstm}. The ConvLSTM block propagates previous feature information, with its output subsequently convolved and flattened into the displacement feature vector $\mathbf{f}_{evt_j}$. 
In the displacement LSTM module, each $\mathbf{f}_{evt_j}$ is concatenated with the hidden displacement feature vector $\mathbf{h}_{j-1}$, leveraging displacement information from preceding frames, and passed through an MLP to obtain a merged displacement vector $\mathbf{m}_{j}$.   
The $\mathbf{h}_{j-1}$ and $\mathbf{m}_{j}$ are then added using a predicted gate weight, yielding the current hidden displacement feature vector $\mathbf{h}_{j}$. 
Finally, the $\mathbf{h}_{j}$ is processed by a linear layer to estimate the displacement $\Delta\hat{\mathbf{p}}$, representing the offset from the query position $\mathbf{p}_{ref}$.

\noindent{\textbf{Uncertainty Predictor.}}
This module predicts the uncertainty associated with the estimated displacement, which is incorporated into the Kalman filter to produce an optimal estimation.
General uncertainty values range from zero to positive infinity, resulting in a distribution that is too broad and unstable for learning. 
To address this, the network is designed to learn a normalized uncertainty value within the range [0, 1].
Specifically, the predictor also takes $\mathbf{F}_{evt_{ref}}$, $\mathbf{F}_{evt_j}$ and $\mathbf{C}_{evt_j}$ as input, concatenates them, and processes them through CNN with five convolution layers to transform the feature patch to an uncertainty feature vector $\mathbf{f}_{uncert_j}$. A linear layer generates two scores representing ``certain" and ``uncertain," followed by a Softmax function to obtain the normalized uncertainty $\hat{\sigma}$, which is subsequently remapped by a parabolic function to range [0, 10], and then repeated and placed on the diagonal of the covariance matrix $\hat{\mathbf{\Sigma}}_{evt_j} \in \mathbb{R}^{2\times2}$. This matrix is fed into the Kalman filter to represent uncertainty.

\subsection{Image Module} \label{Image Frame Relocalization Module}
Event cameras are not good at measuring fine-grained texture information like traditional cameras, which can easily lead to error accumulation and track loss. To enhance long-term tracking performance and minimize cumulative errors, we developed a lightweight image frame relocalization module with vast receptive fields. The design of our module is inspired by recent works on point tracking, such as PIPs\cite{pips}. However, these existing methods operate slowly, limiting overall processing speed despite the high speed of the event module. To address this, we simplified the architecture design, enabling our module to achieve operation speeds exceeding 50 FPS.

\noindent{\textbf{Image Frame Encoder.}}
The image module takes a reference frame $\mathbf{I}_{ref}$ and a target frame $\mathbf{I}_{j}$ input, both images are encoded by a Pyramid Encoder like RAFT\cite{raft} feature map $\mathbf{F}_{img_{ref}}$ and $\mathbf{F}_{img_j}$. 
This encoder effectively reduces computational resources and time while simultaneously increasing the receptive field through convolution layers, which significantly aids in relocalization. 
We apply bilinear sampling to the reference feature map $\mathbf{F}_{img_{ref}}$ at the given reference point $\mathbf{p}_{ref}$ to obtain the reference feature vector $\mathbf{f}_{img_{ref}}$. The feature vector $\mathbf{f}_{img_j}$ is also sampled from $\mathbf{F}_{img_j}$ at last prediction $\mathbf{p}_{j-1}$. 

To evaluate feature similarity explicitly, the feature map $\mathbf{F}_{img_j}$ is processed through several average pool layers to multiple scales and correlated with reference feature vector $\mathbf{f}_{img_{ref}}$, constructing the correlation pyramid $\mathbf{C}_{pym_j}$.
From last predicted position $\mathbf{p}_{j-1}$, we sample patches $\mathbf{P}_{img_j}$ from multiple scale correlation pyramid $\mathbf{C}_{pym_j}$ and concatenate them to the final correlation map $\mathbf{C}_{img_j}$.
This approach facilitates the acquisition of multi-scale information, further broadening the receptive field and ensuring both accuracy and breadth of predictions. 

\noindent{\textbf{Prediction Head.}}
To involves most information, $\mathbf{f}_{img_{ref}}$, $\mathbf{f}_{img_j}$, $\mathbf{C}_{img_j}$, $\mathbf{p}_{j-1}$ and the embedded $\mathbf{p}_{j-1}$ are gathered together as input to a MLP-Mixer\cite{mlp_mixer} inspired from PIPs\cite{pips} to predict displacement $\Delta\hat{\mathbf{p}}$, ``certain" and ``uncertainty" scores. The scores would be processed the same as the event module to calculate $\hat{\mathbf{\Sigma}}_{img_j}$. 

Given the potential distance between the current and actual positions, this procedure can be iteratively performed by extracting $\mathbf{f}_{img_j}$, $\mathbf{C}_{img_j}$ using the newly predicted $\mathbf{p}_{j}$ from $\mathbf{C}_{pym_j}$. This allows for progressively obtaining the most accurate and confident prediction. Finally, the output $\Delta\hat{\mathbf{p}}$ and $\hat{\mathbf{\Sigma}}_{img_j}$ are integrated into the Kalman filter to generate the final prediction.

\subsection{Kalman Filter} \label{Kalman filter}
We employ a Kalman filter to associate and fuse asynchronous data from event and image modality.
The Kalman filter design offers several advantages.
First, it allows measurements from any module at any timestamp, enabling the seamless fusion of multi-modal data arriving at different rates. 
Second, its process is fully differentiable and allows the network to learn to minimize estimation errors by effectively integrating reliable measurements while mitigating the influence of noise(e.g., occlusion) through uncertainty supervision. 
Third, it requires minimal computational resources and maintains low coupling between the event and image modules, enabling modular architectures that better align with the data and support efficient fusion.

In our framework, the Kalman filter observation requires only the position $\Delta\hat{\mathbf{p}}$, covariance matrix $\hat{\mathbf{\Sigma}}$, and timestamp $t$ as inputs. Consequently, measurements can originate from either the event or image module, seamlessly fusing both modules and predicting the filtered final position $\Delta\tilde{\mathbf{p}}$. 
We assume a simple constant velocity model in the Kalman filter by defining state $x$, state covariance matrix $P$. 
\begin{equation}
\begin{aligned}
    x & = (\text{x}, \text{y}, v_{x}, v_{y})^{\text{T}} \\ 
    P & = \left (\begin{array}{cccc}
    \rho_{xx} & 0 & \rho_{xv_{x}} & 0 \\
    0 & \rho_{yy} & 0 & \rho_{yv_{y}} \\ 
    \rho_{v_{x}x} & 0 & \rho_{v_{x}v_{x}} & 0 \\ 
    0 & \rho_{v_{y}y} & 0 & \rho_{v_{y}v_{y}} \\ 
    \end{array}\right) \\
\end{aligned}
\end{equation}
The state $x$ contains the position and velocity of $\text{x}$ and $\text{y}$, and the state covariance matrix $P$ represents the uncertainty of the internal state.
The process in the Kalman filter is traditional\cite{Klmn1961NewRI, Welch1995AnIT}. 
As a new measurement arrives, the Kalman filter performs a prediction step to propagate the internal state and uncertainty to the current timestamp, relying solely on historical information.
The internal state and uncertainty are then updated using the current measurement, incorporating $\Delta\hat{\mathbf{p}}$ and $\hat{\mathbf{\Sigma}}$ from our event or image module.
\begin{equation}
\begin{aligned}
\hat{x}_{k|k-1}, P_{k|k-1} &= \text{Predict}(\hat{x}_{k-1|k-1}, P_{k-1|k-1}), \\
\hat{x}_{k|k}, P_{k|k} &= \text{Update}(z_k, R_k, \hat{x}_{k|k-1}, P_{k|k-1}).
\end{aligned}
\end{equation}
$z$ is the new measurement(our predicted $\Delta\hat{\mathbf{p}}$), $R$ is the measurement noise covariance(our predicted $\hat{\mathbf{\Sigma}}$), and our final prediction $\Delta\tilde{\mathbf{p}}=(\text{x}, \text{y})$ is from the first two elements in the vector $\hat{x}_{k|k}$. 
Throughout the Kalman filter full process, the measurements $\Delta\hat{\mathbf{p}}_k$ are fused with previous state $\hat{x}_{k-1|k-1}$ weighted by $\hat{\mathbf{\Sigma}}$ and $P_{k-1|k-1}$ to get final state $\hat{x}_{k|k}$(i.e., $\Delta\tilde{\mathbf{p}}_k$ and velocity), while simultaneously maintaining velocity $(v_{x}, v_{y})$ and the state covariance matrix $P_{k|k}$.
Both modules are expected to predict high uncertainty in challenging scenarios, such as occlusions, and low uncertainty in confident conditions.
When $\hat{\mathbf{\Sigma}}$ indicates high uncertainty, the Kalman filter relies more on its internal constant velocity model to predict $\Delta\tilde{\mathbf{p}}$. Conversely, when $\hat{\mathbf{\Sigma}}$ indicates low uncertainty, $\Delta\hat{\mathbf{p}}$ is assigned greater weight in determining $\Delta\tilde{\mathbf{p}}$.
Moreover, our predicted $\hat{\mathbf{\Sigma}}$ can be directly supervised through this process, rather than being traditionally handcrafted, significantly improving the quality of the uncertainty. 
This process filters out unreliable predictions and non-uniform noise, ensuring a stable and accurate trajectory with minimal computation. 
More details of operation and definitions in the Kalman filter are provided in the supplementary material.

\begin{figure*}[tb]
  \centering
  \includegraphics[width=0.9\linewidth]{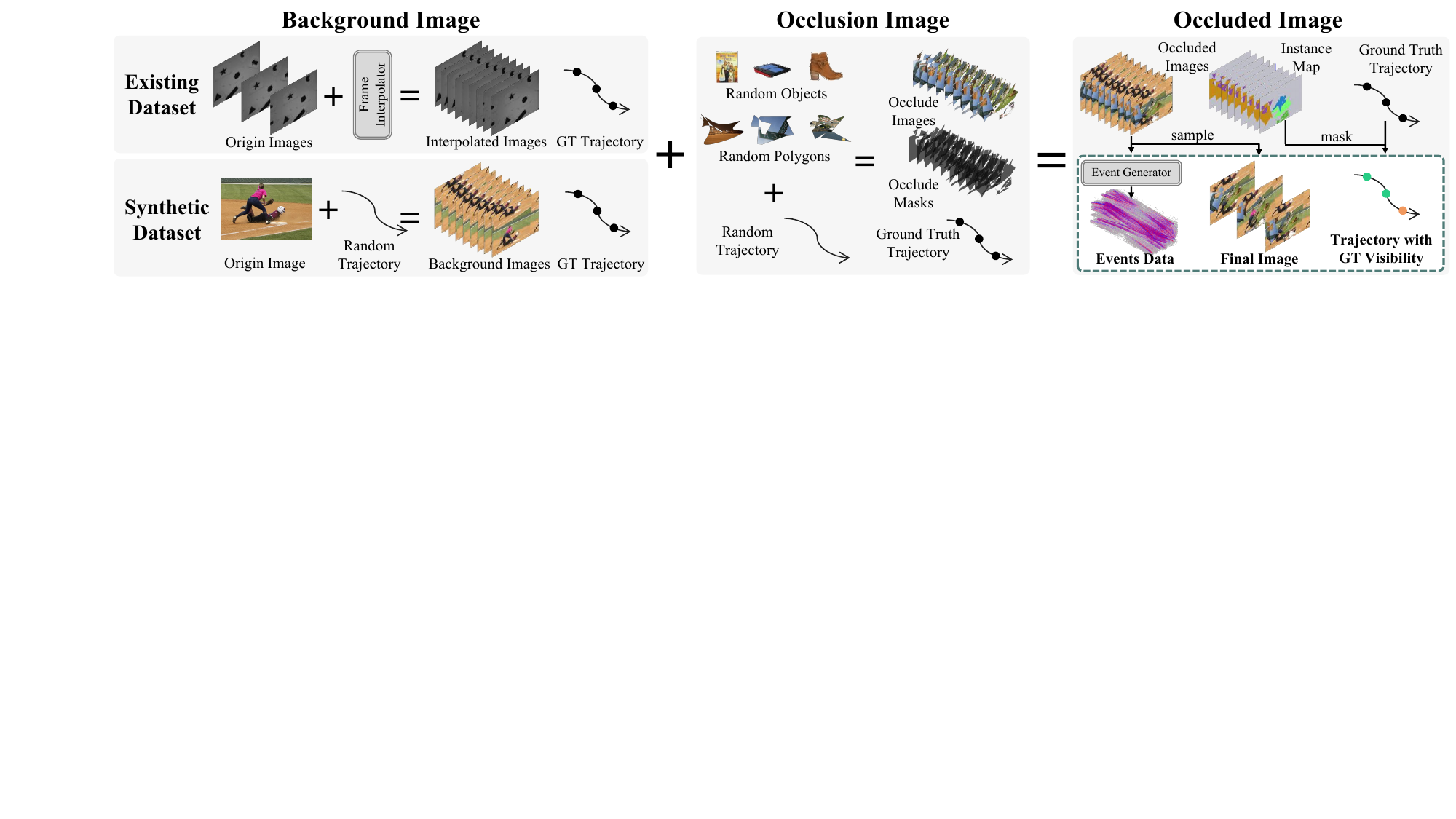}
  \caption{\textbf{Overview of occluded data synthetic pipeline.} The high frame rate occlusion images are overlaid on the high frame rate background images to compose occluded images, which are used to generate synthetic event data.}
  \label{fig:occ}
\end{figure*}

\subsection{Supervision} \label{Supervision}

Joint training of the event and image modules would require excessive computational resources, so we opted to train the two modules separately.
Moreover, we empirically observe that training the entire event module from scratch is unstable and slow, due to cumulative error and the limited receptive field of the event module, which covers only a few dozen pixels.
Therefore, we initially train the event module without the uncertainty predictor and Kalman filter, using only the event displacement loss calculated from the direct predictions $\Delta\hat{\mathbf{p}}$ of the event module. After that, we freeze the encoder and displacement predictor, then train the uncertainty predictor by incorporating the differentiable Kalman filter, using both displacement loss calculated from Kalman filter prediction $\Delta\tilde{\mathbf{p}}$ and event uncertainty loss. The training is conducted on our proposed dataset, MultiTrack (see Sec.~\ref{MultiTrack}). 
While MultiTrack provides a ground truth trajectory for any pixel, we use SuperPoint\cite{superpoint} to extract keypoints as the query points.
In the following, we introduce the two loss functions used to train the event module. The training loss for the image module is similar and detailed in the supplementary material.

\noindent{\textbf{Event Displacement Loss.}}
The displacement loss measures the L1 distance between the predicted displacement $\Delta\hat{\mathbf{p}}$ and ground truth displacement $\Delta{\mathbf{p}}$.
It is important to note that due to error accumulation, the ground truth tracking position may fall outside the search region(i.e., the event patch $\mathbf{P}_{evt_{j}}$). In such cases, we truncate the loss, formulated as follows:

\begin{equation}
\begin{aligned}
    & \mathcal{L}_{\hat{disp}} = \left\{
    \begin{array}{lcl}
    \mid\mid \Delta\hat{\mathbf{p}} - \Delta\mathbf{p} \mid\mid_{1},
    & & \mid\mid \Delta\mathbf{p} \mid\mid_{1} < r\\
    0.& & \text{else}
    \end{array} \right.
\end{aligned} 
\label{Eq: event displacement loss}
\end{equation}

Additionally, we perform on-the-fly augmentation\cite{deepev}, applying random affine transformations to the event patch.

\noindent{\textbf{Event Uncertainty Loss.}}
The uncertainty can be indirectly supervised through displacement loss on $\Delta\tilde{\mathbf{p}}$ by incorporating the Kalman filter. To boost convergence, we also apply visibility loss as a proxy directly on the predicted normalized uncertainty, as visibility probability is strongly correlated with uncertainty. Specifically, we use cross-entropy loss:
\begin{equation}
\begin{aligned}
    \mathcal{L}_{vis} = \text{CrossEntropy}(1-\hat{\sigma}, g),
\end{aligned} \label{Eq: event uncert loss}
\end{equation}
where $g$ is the ground truth visibility. 
We use these two loss functions to train the uncertainty predictor: $\mathcal{L}_{event} = \mathcal{L}_{\tilde{disp}} + w_{1}\mathcal{L}_{vis}$, where $w_{1} = 2$.

\section{MultiTrack, EC-occ and EDS-occ} \label{MultiTrack, EC-occ and EDS-occ}
As manually labeling visibility ground truth is highly challenging, we introduce a data synthesis pipeline, see Fig.~\ref{fig:occ}. Our pipeline can synthesize color images, events, occluded tracks, and visibility status with adjustable parameters, or generate these data from existing datasets, which require only raw color images and specified pixel trajectories.

\subsection{MultiTrack}\label{MultiTrack}
Following MultiFlow\cite{multiflow}, our generated dataset is based on the premise that the background exhibits regular motion, whereas certain foreground objects present rapid motion and occlusion.
The background images are selected from the Flickr 30k dataset\cite{flickr30k}, while the foreground images are of two types to provide more comprehensive features and occlusions. The first type consists of objects with transparent backgrounds from Google Scanned Objects\cite{googlescanned}, the other type has random polygons or smoothed shapes with random images hatched in these shapes. The shapes are inspired by AutoFlow\cite{autoflow}, and the hatched images are from COCO2014\cite{coco}.
With random translation, rotation, and scale of each image, the foreground images are overlaid together on background images, producing synthetic high frame rate(1000 Hz) images, instance maps, and dense complex tracks with visibility. 
The high frame rate images are taken as input to DVS-Voltmeter\cite{Lin2022DVSVoltmeterSP} to generate synthetic event data while they are also sampled in a normal frame rate to get a color image. By using this pipeline, we generate MultiTrack as our training data.

\subsection{EC-occ, EDS-occ}

When applying occlusion on an existing dataset, we first interpolate the images using FILM\cite{reda2022film}, which iterates 4 times, inserting 15 frames for each color frame interval(final 384 Hz for EC\cite{ec} and 1200 Hz for EDS\cite{eds}). With interpolated high frame rate background images, the same procedures are applied to create occlusion on the existing dataset, obtaining occluded event data from DVS-Voltmeter\cite{Lin2022DVSVoltmeterSP} and visibility of given tracks from the instance map.
By applying this pipeline, EC\cite{ec} and EDS\cite{eds} are occluded and processed to EC-occ and EDS-occ.

\section{Experiment}

\noindent{\textbf{Dataset.}}
We test our method in a commonly used real captured dataset, Event Camera dataset(EC)\cite{ec}, and Event-aided Direct Sparse Odometry dataset(EDS)\cite{eds}. EC provides $240 \times 180$ pixels 24Hz APS frames and events, which is recorded by a DAVIS240C camera\cite{6889103} while EDS contains $640 \times 480$ pixels 75Hz image data and event data captured with a beam splitter setup\cite{eds}. The events from EC and EDS are pre-processed to SBT-Max\cite{sbt} with intervals of 10ms and 5ms, so that the input event data are in frequency of 100Hz and 200Hz, respectively. These datasets do not originally offer the ground truth for feature tracking, and Messikommer et al.\cite{deepev} extend it by calculating the ground truth through KLT\cite{klt} tracking and triangulation. We also test in our synthetic dataset with occlusion, EC-occ, and EDS-occ introduced at Sec.~\ref{MultiTrack, EC-occ and EDS-occ}. Another widely used event-based dataset, DSEC\cite{dsec, eraft}, is also used in qualitative comparison to evaluate performance on large outdoor scenes.

\noindent{\textbf{Baselines.}}
We conducted experiments against the state-of-the-art method Deep-EV-Tracker\cite{deepev}, which tracks event frames using grayscale image references with smaller patches. We also obtained experimental data from the concurrent method Deep-Ev-Tracker(E2VID)\cite{deepev2}.
We also compare with EKLT\cite{eklt}, which has a similar pipeline with Deep-EV-Tracker but without using a neural network. Additionally, HASTE\cite{haste}, which demands no image but uses pure event data, is included. Additionally, to demonstrate our novel multi-modality combination technique, we follow Deep-EV-Tracker\cite{deepev}, combining it with KLT\cite{klt}, a well-established color frame tracker, by replacing the initial position with predictions from another module. Our fusion method without a Kalman filter follows the same approach. We also evaluate FF-KDT\cite{ffkdt}; however, it produces estimates only at image timestamps, whereas Deep-Ev-Tracker\cite{deepev} and our method generate significantly more predictions at event timestamps.

\noindent{\textbf{Metric.}}
The EC\cite{ec} and EDS\cite{eds} only provide ground truth trajectories with 24Hz and 75Hz. To evaluate the high-rate event predictions with 100Hz and 200Hz, we interpolate the event predictions to continuous trajectories and sample points on those timestamps that have ground truth data following Messikommer et al.\cite{deepev}.
We evaluate feature age(FA), defined as the duration for which a feature exceeds a specified distance(1, 2, 3 ..., 32) from the ground truth, and expect feature age(Exp FA or Expect FA), which is calculated as the product of feature age and the ratio of stable tracks—those maintaining errors below the threshold across all frames. These two metrics are common in event-based feature tracking benchmarks\cite{deepev}. We also evaluate accuracy through $\delta_{avg}$\cite{tapvid}, which is the average fraction of points tracked within 1, 2, 4, 8, and 16 pixels, also widely used in point tracking benchmarks\cite{pointodyssey, cotracker}. 

\noindent{\textbf{Implementation Details.}}
As mentioned in Sec.~\ref{Supervision}, our event module is pre-trained on the MultiTrack dataset on 30000 feature tracks from 2000 sequences employing a continual learning approach\cite{deepev}. We use ADAM\cite{adam} optimizer with a learning rate of $1 \times 10^{-4}$ and 140000 training steps in total with batch size 32. To initialize training, the input sequences are clipped to 4 frames, which would increase to 12 and 23 after 80000 and 120000 steps\cite{deepev}. Then we train our uncertainty predictor on the same dataset with the same hyperparameters for 80000 steps. 
At the same time, we train our image module on MultiTrack, which is generated for image training with 10000 sequences using ADAMW\cite{adamw} with a learning rate of $5 \times 10^{-4}$ and batch size 16. 
The sequence length is fixed to 10 in all 50000 training steps. We run both modules' supervision on two NVIDIA RTX3090 24GB GPUS for 48 hours and experiments on one NVIDIA RTX3090 GPU and Intel(R) Xeon(R) Gold 6139M CPU.

\subsection{Experiments Results}

\begin{table}[tb]
\centering
\scriptsize
\setlength\tabcolsep{1.8pt}
\renewcommand\arraystretch{1}
\begin{tabular}{c||l||ccc||ccc}
\toprule
\multirow{2}{*}{Data} & \multirow{2}{*}{Methods} & \multicolumn{3}{c||}{EC} & \multicolumn{3}{c}{EDS} \\ 
 & & FA$\uparrow$ & Exp FA$\uparrow$ & $N_{p}$$\uparrow$ & FA$\uparrow$ & Exp FA$\uparrow$ & $N_{p}$$\uparrow$ \\ \bottomrule
 & HASTE\cite{haste}                & 0.442          & 0.427   &   -    & 0.096          & 0.063     &  -   \\
& EKLT\cite{eklt}             & 0.811 &0.775 &-& 0.325& 0.205& -\\

E& D-Tracker\cite{deepev}   & 0.795          & 0.787  &100        & 0.549          & 0.451    &200      \\
& D-Tracker(E2VID)\cite{deepev2}   & 0.793          & 0.781   &   100    & \textbf{0.579}          & \textbf{0.482}  &    200    \\
& \textbf{Ours(E)}           & \underline{0.833}          & \underline{0.819}  &100        & 0.568          & 0.474 &200      \\
& \textbf{Ours(E w. K)}      & \textbf{0.835} & \textbf{0.820} & 100&\underline{0.569} & \underline{0.475} &200\\ \midrule

\multirow{4}{*}{\shortstack{E\\+\\I}} & FF-KDT\cite{ffkdt} & 0.852*          & 0.846*   &24       & 0.526*          & 0.431*  &75   \\
&D-Tracker\cite{deepev}+KLT\cite{klt} & 0.735          & 0.730   &124       & 0.594          & 0.503     &275     \\
&\textbf{Ours(E+I)}  & \underline{0.786}   & \underline{0.780}  &124  & \underline{0.634} & \underline{0.532}&275  \\
&\textbf{Ours(E+I w. K)}  & \textbf{0.851} & \textbf{0.845} & 124&\textbf{0.653} & \textbf{0.550}&275\\ \midrule
\end{tabular}
\caption{\textbf{Performance evaluation on EC and EDS.} ``D-tracker'' denotes Deep-EV-Tracker\cite{deepev}. ``E'' denotes event, ``I'' denotes image and ``K'' denotes Kalman. $N_{p}$ denotes the number of predictions per second of data. HASTE\cite{haste} and EKLT\cite{eklt} operate on raw event data without frames, resulting in an unstable number of predictions. * FF-KDT\cite{ffkdt} results are not compared due to their limited prediction number.}
\label{tab:eceds}
\end{table}
\noindent{\textbf{Real Data Experiment.}}
We compare our method with other state-of-the-art methods in terms of the input signals, i.e., event-only, and both event and image, in Tab.~\ref{tab:eceds}. Note that even though both EC\cite{ec} and EDS\cite{eds} only contain static scenarios scenes with almost no occluded trajectories, which are rather simple and do not fully reveal the superiority of our learnable Kalman filter, our method still presents state-of-the-art performance on both tracks and obtains high-frequency predictions when fusing the event and image signals. Due to their simplicity, a naive LSTM within the event module can capture contextual information across frames implicitly, and our proposed learnable Kalman filter does not bring a prominent performance gain.

\begin{table*}[tb]
\centering
\scriptsize
\setlength\tabcolsep{5.5pt}
\renewcommand{\arraystretch}{1}
\begin{tabular}{c||c||ccccc||ccccc}
\toprule
Data & Metric & \multicolumn{5}{c||}{EC-occ} & \multicolumn{5}{c}{EDS-occ} \\ \midrule  
\multicolumn{1}{c}{}         &     \multicolumn{1}{c}{}    & \multicolumn{1}{c}{FF-KDT\cite{ffkdt}}     & \multicolumn{1}{c}{D-tracker\cite{deepev}}                           & \multicolumn{1}{c}{\textbf{Ours(E)}}   & \multicolumn{1}{c}{\textbf{Ours(E+K)}}   & \multicolumn{1}{c}{Incr(\%)$\uparrow$}  & \multicolumn{1}{c}{FF-KDT\cite{ffkdt}}    & \multicolumn{1}{c}{D-tracker\cite{deepev}}                           & \multicolumn{1}{c}{\textbf{Ours(E)}}   & \multicolumn{1}{c}{\textbf{Ours(E+K)}}   & \multicolumn{1}{c}{Incr(\%)$\uparrow$} \\
  \midrule
    & $\delta_{}^{vis}\uparrow$& -    & 26.8       & 29.3     & \textbf{30.5}           & 3.92& -    & 31.3  & 35.1   & \textbf{36.7}                            & 4.50                         \\
E& $\delta_{}^{occ}\uparrow$& -    & 20.2                                                          & 23.1                          & \textbf{24.1}                            & \textbf{4.41}                        & -    & 21.3                                                          & 24.6                          & \textbf{25.9}                            & \textbf{5.05}                         \\
& $\delta_{}^{all}\uparrow$& -    & 26.3                                                          & 28.7                          & \textbf{29.8}                            & 3.95                        & -    & 30.9                                                          & 34.6                          & \textbf{36.2}                            & 4.51                         \\ \midrule

\multicolumn{1}{c}{}         &     \multicolumn{1}{c}{}    & \multicolumn{1}{c}{\multirow{2}{*}{FF-KDT\cite{ffkdt}}} &\multicolumn{1}{c}{D-tracker\cite{deepev}} & \multicolumn{1}{c}{\textbf{Ours}} & \multicolumn{1}{c}{\textbf{Ours}} & \multicolumn{1}{c}{\multirow{2}{*}{Incr(\%)$\uparrow$}} & \multicolumn{1}{c}{\multirow{2}{*}{FF-KDT\cite{ffkdt}}}  &\multicolumn{1}{c}{D-tracker\cite{deepev}} & \multicolumn{1}{c}{\textbf{Ours}} & \multicolumn{1}{c}{\textbf{Ours}} & \multicolumn{1}{c}{\multirow{2}{*}{Incr(\%)$\uparrow$}} \\ 
   \multicolumn{1}{c}{}         &     \multicolumn{1}{c}{}    &     \multicolumn{1}{c}{}    & \multicolumn{1}{c}{+ KLT\cite{klt}} & \multicolumn{1}{c}{\textbf{(E+I)}} & \multicolumn{1}{c}{\textbf{(E+I w. K)}} &\multicolumn{1}{c}{}     &     \multicolumn{1}{c}{}   & \multicolumn{1}{c}{+ KLT\cite{klt}} & \multicolumn{1}{c}{\textbf{(E+I)}} & \multicolumn{1}{c}{\textbf{(E+I w. K)}} &  \multicolumn{1}{c}{}  \\ \midrule
& $\delta_{}^{vis}\uparrow$&28.2   & 32.1                                                          & 37.2                          & \textbf{44.5}                            & 19.73                & 24.0          & 35.0                                                          & 43.1                          & \textbf{52.0}                            & 20.64                        \\
E+I& $\delta_{}^{occ}\uparrow$ & 14.5   & 5.8                                                           & 11.4                          & \textbf{28.5}                            & \textbf{149.06}                 & 11.5       & 10.7                                                          & 18.6                          & \textbf{26.8}                            & \textbf{43.61}                      \\
& $\delta_{}^{all}\uparrow$& 27.1   & 30.3                                                          & 35.5                          & \textbf{43.4}                            & 22.32                       & 23.4   & 33.7                                                          & 41.6                          & \textbf{50.6}                            & 21.51     \\  \midrule                  

\end{tabular}
\caption{\textbf{Performance increase from module without Kalman filter to module with Kalman filter.} 
``Incr'' denotes the performance increase from our module without a Kalman filter to those with a Kalman filter.}

\label{tab:Incr}
\end{table*}

\noindent{\textbf{Occlusion Data Experiment.}}
To better demonstrate the superiority of the Kalman filter, experiments are conducted on more challenging datasets with occlusions, EC-occ, and EDS-occ, see Tab.~\ref{tab:Incr}. 
The improved performance of modules incorporating the Kalman filter demonstrates its effectiveness in stabilizing tracking. This is particularly evident in occluded scenarios, where occluded points exhibit the most significant performance gains, confirming that the Kalman filter mitigates instabilities and corrects erroneous measurements caused by noise.

Both experiments demonstrate that naively combining the two modules by replacing initial points does not yield satisfactory results and may even degrade performance due to the compounding limitations of both.
However, with the aid of the Kalman filter, significant improvements are observed, demonstrating its effectiveness in combining and optimizing asynchronous multimodal predictions.
In detail, uncertainty increases in challenging scenarios for event or traditional cameras, such as extreme lighting or motion blur for traditional cameras and low-texture environments for event cameras. In these cases, the Kalman filter mitigates the impact of biased predictions by accounting for high uncertainty. Conversely, when cameras operate in optimal conditions, uncertainty remains low, allowing predictions to have a greater influence on the Kalman filter’s output. This process harnesses the strengths of both event and image cameras while mitigating their limitations, resulting in a more balanced and accurate final prediction.

\begin{figure}[tb]
  \centering
  \includegraphics[width=1\linewidth]{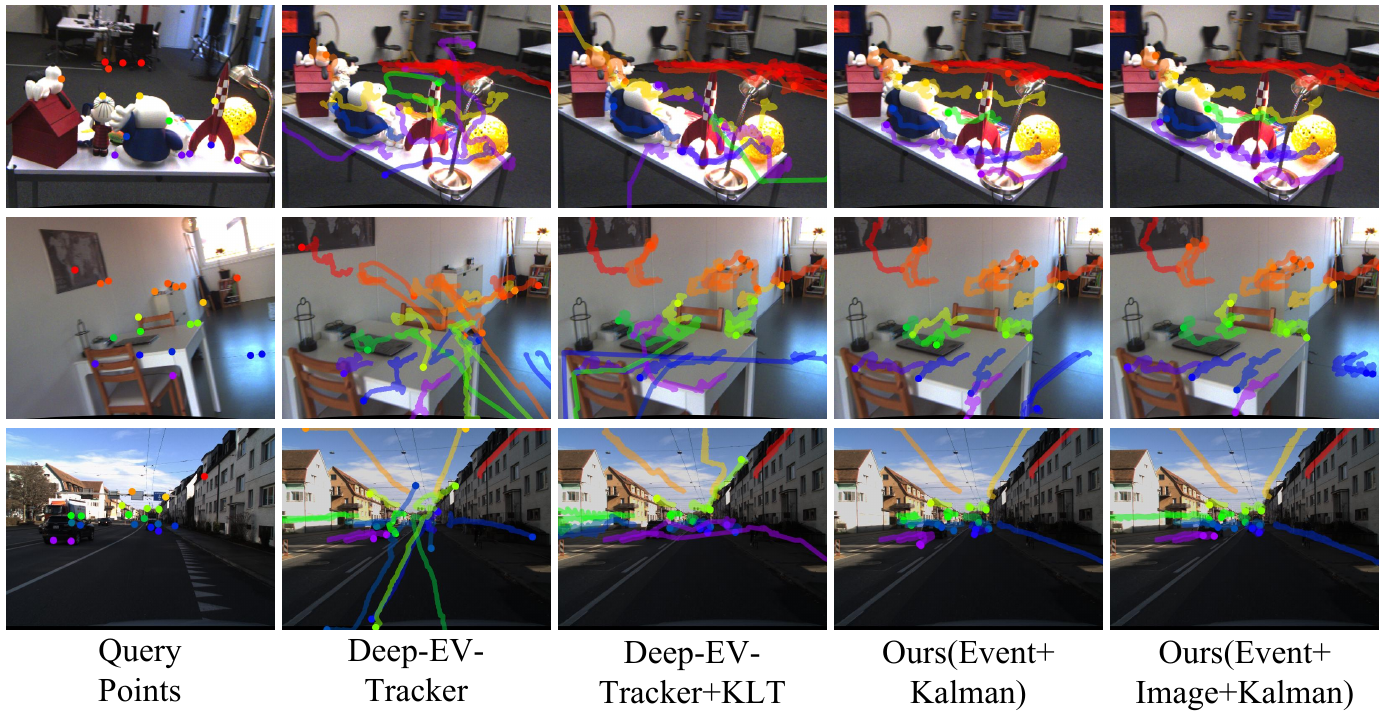}
  \caption{\textbf{Qualitative comparison on EDS\cite{eds}(top two rows) and DSEC\cite{dsec}(bottom row).}}
  \label{fig:result}
\end{figure}

\noindent{\textbf{Qualitative Comparison.}} 
We visualize the tracks estimated by our methods and other methods in Fig.~\ref{fig:result}.
As shown, all event-based methods exhibit some degree of error; however, our event module produces fewer lost tracks compared to the Deep-EV-Tracker\cite{deepev}, indicating greater robustness. 
With the integration of the image module, our method achieves improved performance, demonstrating that the image module is effective in enhancing relocalization and accuracy. 
When comparing the multi-modal results, a simple fusion approach, such as replacing initial points, results in a higher number of lost tracks, whereas Kalman fusion proves to be more stable and reliable.

\noindent{\textbf{Runtime Analysis.}}
Our event module takes less than 9 ms to track one event patch across one preprocessed event frame on EC\cite{ec} and EDS\cite{eds} using an Nvidia RTX3090. The introduction of the Kalman filter adds less than 1 ms to the budget. 
The preprocessing of event data takes 3ms on EC ($240\times180$ resolution, 10ms frame interval) and 4ms on EDS ($640\times480$ resolution, 5ms frame interval). It is executed on the CPU and can run in parallel with the tracking model. Deep-Ev-Tracker\cite{deepev} requires identical preprocessing with ours, while Deep-Ev-Tracker(E2VID)\cite{deepev2} processes each frame in 153ms\cite{deepev2}, and FF-KDT\cite{ffkdt} takes 13ms on EC and 79ms on EDS. As a result, our integration of the event module and Kalman filter enables precise predictions at frequencies exceeding 100 FPS, while Deep-Ev-Tracker\cite{deepev} only achieves 50 FPS. 
Additionally, our image module can operate efficiently with the Kalman filter at frequencies above 50 FPS, making it efficient enough for low-frequency image relocalization.

If the image module and event module run sequentially, with the image module operating at intervals of 0.05s and producing 20 predictions per second, it accounts for 0.4s of the runtime. The event module then uses the remaining 0.6s to produce at least 60 predictions. Consequently, the entire model generates 80 predictions per second, while FF-KDT\cite{ffkdt} achieves only 7 FPS.

\begin{table}[tb]
\centering
\scriptsize
\setlength\tabcolsep{1.5pt}
\renewcommand\arraystretch{1.1}
\begin{tabular}{l||l||cc||cc||c}
\toprule
\multicolumn{7}{c}{Expect FA$\uparrow$} \\ \midrule
Experiment & Method & EC & EC-occ & EDS & EDS-occ & Paras \\ \bottomrule
Feature Pyramid        & \underline{On}                                 & \textbf{0.819} & \textbf{0.522} &\textbf{0.474} & \textbf{0.343} & 33.1M \\
(Ours Event) & Off                                      & 0.789 & 0.439 & 0.395 & 0.283 & 29.3M \\\midrule
LSTM Module            & \underline{Feature+Displace}                 & \textbf{0.819} & \textbf{0.522} & \textbf{0.474} & \textbf{0.343} & 33.1M \\
(Ours Event)  & Feature                                    & 0.794  & 0.473  & 0.428  & 0.337  & 32.7M    \\
                       & Displacement                             & 0.617  & 0.332  & 0.282  & 0.186  & 31.9M    \\
                       & Off                                      & 0.568 & 0.338 & 0.294 & 0.225 & 31.6M \\\midrule
Correlation Vector     & \underline{Feature Map Center}                 & \textbf{0.819} & \textbf{0.522} & \textbf{0.474} & \textbf{0.343} & 33.1M \\
(Ours Event) & U-Net Bottleneck                         & 0.792 & 0.472 & 0.441 & 0.330 & 35.7M    \\\midrule

\end{tabular}
\caption{\textbf{Ablation experiments.} The ablated modules are labeled under the experiments. Settings used in our final model are underlined.}

\label{tab:ablation}
\end{table}
\subsection{Ablations}
To evaluate the distinct contributions of each integrated network component on tracking accuracy,
we conducted a series of ablation studies as outlined in Tab.~\ref{tab:ablation}. 

\noindent{\textbf{Feature Pyramid.}}
The feature pyramid is used to take both the perception field and accuracy into account. We conducted experiments replacing the pyramid feature encoder with a simple patch encoder\cite{deepev}, which resulted in reduced performance due to the limited information from the encoder. This outcome suggests that the Feature Pyramid enhances tracking accuracy and effectively handles occlusions. 

\noindent{\textbf{LSTM Module.}}
Our event module utilizes two LSTM modules, specifically designed to convey previous feature and displacement information. We evaluated variants with one LSTM module removed and another without any LSTM modules. The results indicate that removing either LSTM module results in a decline in performance, underscoring the contributions of both modules.

\noindent{\textbf{Correlation Vector.}}
The Deep-EV-Tracker\cite{deepev} uses the U-Net bottleneck vector as the reference vector for correlating with the target feature map. However, we propose that the reference vector should be derived from the reference feature map to maintain symmetry. Our experiment indicates that using the vector from the reference feature map yields better performance.

\section{Conclusion}
In this paper, we presented BlinkTrack, a novel framework for high-frame-rate feature tracking that leverages the strengths of both event data and images. By integrating a differentiable Kalman filter within a learning-based architecture, BlinkTrack effectively addresses the challenges of asynchronous data fusion and improves tracking performance under multimodal data with noise(e.g., occlusion). Our extensive experiments, supported by newly generated and augmented datasets, show that BlinkTrack significantly outperforms existing methods in terms of robustness and speed, achieving state-of-the-art performance while running at over 80 FPS with multi-modality data and 100 FPS with preprocessed event data. 
These results underscore the potential of our approach for advanced feature tracking applications, establishing a new benchmark for future research in this field.
\textbf{Limitations.}
One limitation of the current performance may be the network capacity and the receptive field. In future work, we aim to investigate strategies for scaling up the network parameters while maintaining efficiency(e.g., achieving over 100 FPS).

\noindent{\textbf{Acknowledgements:}}
This work was partially supported by the NSF of China(No.62425209).


\clearpage

\setcounter{section}{0}
\setcounter{table}{0}
\setcounter{figure}{0}
\setcounter{equation}{0}
\renewcommand\thesection{\Alph{section}}
\renewcommand\thetable{\Alph{table}}
\renewcommand\thefigure{\Alph{figure}}
\renewcommand\theequation{\Alph{equation}}

\maketitlesupplementary

In this supplementary document, we first introduce more details about BlinkTrack in Sec.~\ref{More Details about BlinkTrack}, including an explanation of event representation and pre-processing in Sec.~\ref{Event Representation and Pre-processing}, detail of the image module supervision in Sec.~\ref{Details of the Image Module and Its Supervision}, model definition in Kalman filter in Sec.~\ref{Kalman Filter Definition}, and more implementation detail in Sec.~\ref{More Implementation Details}.
Then we present more experiment results in Sec.~\ref{More Experiment Results}, including quantitative comparisons in Sec.~\ref{More Quantitative Comparisons}, qualitative comparisons in Sec.~\ref{More Qualitative Comparison}, uncertainty visualization in Sec.~\ref{Uncertainty Visualization}, runtime comparison between our method and other methods in Sec.~\ref{More Runtime Analysis}, and discussion on reference frame in Sec.~\ref{Event Module Reference Frame Experiment}.
At last, we add more details about our dataset in Sec.~\ref{More Details about MultiTrack, EC-occ and EDS-occ}, including displacement distribution analysis in Sec.~\ref{Displacement Distribution of MultiFlow and MultiTrack}, data quality experiment in Sec.~\ref{Data Quality of MultiFlow and MultiTrack}, synthetic data quality experiment in Sec.~\ref{Synthetic Data Quality of EC-occ and EDS-occ}, some visualization examples in Sec.~\ref{Visualization Examples of MultiTrack, EC-occ and EDS-occ}, and license explanation in Sec.~\ref{Dataset License}.

\section{More Details about BlinkTrack} \label{More Details about BlinkTrack}

\subsection{Event Representation and Pre-processing} \label{Event Representation and Pre-processing}

A single event $\mathbf{e}_{k} = \{x_{k}, y_{k}, t_{k}, p_{k}\}$ is triggered when pixel $(x_{k}, y_{k})$ have brightness change indicated by $p_{k} \in \{-1, +1\}$ at time $t_{k}$\cite{wu2023flytracker}. The event stream $E = \{\mathbf{e}_{k}\}_{k=1}^{n}$ is composed of a large amount of discrete single event $e_{k}$. The frame-based neural network cannot directly process such data representations; therefore, extracting meaningful information from discrete events and converting them into spatially and temporally aligned frames is essential. According to the representation experiments\cite{deepev}, we choose SBT-Max\cite{sbt}. The representation we use has 5 temporal bins for positive and negative polarity(i.e., the sign of the brightness change), respectively, and the streaming events in between the times are merged into those bins. 

We denote the event frame interval as $\Delta{t}$ and we want to extract event frame $\mathbf{E}_{j}$ from events $E_{j} = \{\mathbf{e}_{k}\}_{k=1}^{n_{evt}}$ between $t_{j} - \Delta{t}$ and $t_{j}$ with $n_{bin}$ bins which means interval for each bin is $\frac{\Delta{t}}{n_{bin}}$, for example the time interval of first bin is $(t_{j} - \Delta{t}, t_{j} - \Delta{t} + \frac{\Delta{t}}{n_{bin}}]$. Each pair of positive and negative channels is formed by the maximal timestamp of positive or negative polarity, respectively, on a single pixel among events from $(t, t + \frac{\Delta{t}}{n_{bin}}]$. If there is no event with the expected polarity between this interval for some pixels, the number of the expected polarity in this bin for these pixels would be set to $0$, see Eq.~\ref{Eq: Event Representation}.

\begin{equation}
\begin{aligned}
& \mathbf{E}_{j}(u, v, 2d) = \left\{
\begin{array}{lcl}
max(t_{k} - t\mid p_{k}=-1, u=x_{k}, v=y_{k}) & & {}\\
0, \text{if there is no such event}
\end{array} \right. \\
& \mathbf{E}_{j}(u, v, 2d+1) = \left\{
\begin{array}{lcl}
max(t_{k} - t\mid p_{k}=1, u=x_{k}, v=y_{k}) & & {}\\
0, \text{if there is no such event}
\end{array} \right. \\
& d = \{0, \dots, n_{bin}\}, k = \{1, \dots, n_{evt}\}
\end{aligned}
\label{Eq: Event Representation}
\end{equation}

\subsection{Details of the Image Module Supervision} \label{Details of the Image Module and Its Supervision}
The image module is also trained on our MultiTrack dataset to align with the event module, employing similar supervision. Due to its lightweight architecture and expansive receptive field, the Kalman filter and uncertainty supervision are applied from the outset, calculating displacement loss based on both the module direct predictions $\Delta\hat{\mathbf{p}}$ and Kalman filter predictions $\Delta\tilde{\mathbf{p}}$. Given the importance of the Kalman filter's predictions in supervising both accuracy and uncertainty, the other losses are weighted by $w_{2} = \frac{1}{2}$ and $w_{3} = \frac{1}{2}$.

\begin{equation}
\begin{aligned}
    & \mathcal{L}_{\hat{disp}} = \mid\mid \Delta\hat{\mathbf{p}} - \Delta\mathbf{p} \mid\mid_{1} \\
    & \mathcal{L}_{\tilde{disp}} = \mid\mid \Delta\tilde{\mathbf{p}} - \Delta\mathbf{p} \mid\mid_{1} \\
    & \mathcal{L}_{image} = \mathcal{L}_{\tilde{disp}} + w_{2}\mathcal{L}_{\hat{disp}} + w_{3}\mathcal{L}_{uncert}
\end{aligned} \label{Eq: image loss}
\end{equation}

Since the Kalman filter and uncertainty module are activated from the beginning, stable supervision is achieved by selecting only points visible in the initial target frame, enabling the Kalman filter to initialize velocity rather than maintain a random initial velocity. To accelerate convergence and improve performance on challenging points, supervision points are sampled based on the weight of their prediction loss, with points exhibiting higher loss being more likely to receive increased supervision.

\subsection{Kalman Filter Definition} \label{Kalman Filter Definition}
The Kalman filter\cite{Klmn1961NewRI, Welch1995AnIT} comprises two fundamental steps: prediction and update. The prediction step estimates the state and its uncertainty at the next time step based on the current state and process model, allowing the internal state to evolve even when no new measurement is available or the measurement is highly uncertain. The prediction step is formulated as:
\begin{equation}
\begin{aligned}
\hat{x}_{k|k-1} &= F \hat{x}_{k-1|k-1}, \\
P_{k|k-1} &= F P_{k-1|k-1} F^T + Q,
\end{aligned}
\label{eq:kalman1}
\end{equation}
where $x$ is the state estimate, $P$ is the state covariance, $F$ is the state transition model, and  $Q$ is the process noise covariance.
The update step incorporates new measurements to refine the state estimate and reduce uncertainty, which is formulated as:
\begin{equation}
\begin{aligned}
y_k &= z_k - H \hat{x}_{k|k-1}, \\
S_k &= H P_{k|k-1} H^T + R, \\
K_k &= P_{k|k-1} H^T S_k^{-1}, \\
\hat{x}_{k|k} &= \hat{x}_{k|k-1} + K_k y_k, \\
P_{k|k} &= (I - K_k H) P_{k|k-1},
\end{aligned}
\label{eq:kalman2}
\end{equation}
where $y$ is the measurement residual, $z$ is the new measurement(our predicted $\Delta\hat{\mathbf{p}}$), $S$ is the residual covariance, $R$ is the measurement noise covariance(our predicted $\hat{\mathbf{\Sigma}}$),  $H$ is the observation model, $K$ is the Kalman gain, and $I$ is the identity matrix. Our final prediction $\Delta\tilde{\mathbf{p}}$ is from the first two elements in the vector $\hat{x}_{k|k}$.

The simple constant velocity model is defined in our Kalman filter. The state $x$, covariance matrix $P$, observation model $H$, state-transition model $F$, and the covariance of the process noise $Q$ are defined as follows.

\begin{equation}
\begin{aligned}
    x & = (\text{x}, \text{y}, v_{x}, v_{y})^{\text{T}} \\ 
    P & = \left (\begin{array}{cccc}
    \rho_{xx} & 0 & \rho_{xv_{x}} & 0 \\
    0 & \rho_{yy} & 0 & \rho_{yv_{y}} \\ 
    \rho_{v_{x}x} & 0 & \rho_{v_{x}v_{x}} & 0 \\ 
    0 & \rho_{v_{y}y} & 0 & \rho_{v_{y}v_{y}} \\ 
    \end{array}\right) \\
    H & = \left (\begin{array}{cccc}
    1 & 0 & 0 & 0 \\
    0 & 1 & 0 & 0
    \end{array}\right) \\
    F & = \left (\begin{array}{cccc}
    1 & 0 & \Delta t & 0 \\
    0 & 1 & 0 & \Delta t \\ 
    0 & 0 & 1 & 0 \\ 
    0 & 0 & 0 & 1 \\ 
    \end{array}\right) \\
    Q & = \left (\begin{array}{cccc}
    \frac{\Delta t^{4}}{4} & 0 & \frac{\Delta t^{3}}{2} & 0 \\
    0 & \frac{\Delta t^{4}}{4} & 0 & \frac{\Delta t^{3}}{2} \\ 
    \frac{\Delta t^{3}}{2} & 0 & \Delta t^{2} & 0 \\ 
    0 & \frac{\Delta t^{3}}{2} & 0 & \Delta t^{2} \\ 
    \end{array}\right)
\end{aligned}
\label{eq:kalman3}
\end{equation}

To explicitly construct the motion model, the state $x$ includes variables of position and velocity of both $\text{x}$ and $\text{y}$. The elements in the covariance matrix $P$ represent the correlation coefficients of these variables, where it is assumed that no correlation exists between $\text{x}$ and $\text{y}$. We initialize state $x$ with $\mathbf{0}$ and covariance matrix $P$ with the Identity matrix, indicating the point is stationary initially with medium confidence. The observation model $H$ specifies that the observations from our modules correspond to the first two elements of the state $x$. The state-transition model $F$ follows a constant velocity assumption, given by $\text{x}_{k} = \text{x}_{k-1} +  v_x\cdot\Delta t $, where $\Delta t$ denotes the time interval between the current and the previous observations. The process noise $Q$ reflects increasing uncertainty in the Kalman state as the duration without observations grows.

In our implementation, when a measurement, including displacement $\Delta\hat{\mathbf{p}}$ and covariance matrix $\hat{\mathbf{\Sigma}}$, arrives at timestamp $t$, the Kalman filter first performs a prediction step up to $t$, generating the predicted state $x$ and state covariance $P$. The state covariance represents the uncertainty of the state, which increases with the time interval since the last prediction. The longer the interval, the higher the uncertainty. Subsequently, all measurements ($\Delta\hat{\mathbf{p}}$, $\hat{\mathbf{\Sigma}}$), along with predicted $x$ and $P$, are processed in the update step, which estimates the final predicted position $\Delta\tilde{\mathbf{p}}$ and updates the state $x$ and state covariance $P$ within the Kalman filter.

We adopt the simplest Kalman filter to validate the effectiveness of our idea. Advanced variants (e.g., the Extended Kalman Filter) can be readily implemented by revising Eq.~\ref{eq:kalman1}, Eq.~\ref{eq:kalman2}, and Eq.~\ref{eq:kalman3}.

\subsection{More Implementation Details} \label{More Implementation Details}
\noindent{\textbf{Event Module.}}
As mentioned in the paper, our extracted patch has size $P_{evt} = 62$. The patches are encoded to reference and event feature maps $\mathbf{F}_{evt}$ and correlation maps $\mathbf{C}_{evt}$ with the same dimension $D_{F_{evt}} = 128$. These feature patches are extracted into a pyramid feature map with $L_{evt} = 2$ layers with size $\{62, 31\}$. In the LSTM displacement predictor, the feature map $\mathbf{F}_{evt}$ turns to a feature vector $\mathbf{f}_{evt}$ with dim $D_{f_{evt}} = 256$, which is obtained by merging a 128 dim hidden feature map patch in feature LSTM and a 256 dim hidden feature vector in displacement LSTM. In the uncertainty predictor, five convolution layers are applied to generate the uncertainty feature vector $\mathbf{f}_{uncert_j}$ with dim $D_{uncert} = 128$. We use 3 points to generate a parabola for uncertainty mapping, which maps $\{0, 0.9, 1\}$ to $\{0, 1, 10\}$. Detailed network information is listed in Tab.~\ref{tab:event_model}.

\begin{table}[tb]
\centering
\scriptsize
\setlength\tabcolsep{7.2pt}
\begin{tabular}{clc}
\toprule
                  & Layer       & Data Size                     \\ \bottomrule

\multirow{22}{*}{\shortstack{Pyramid\\Feature\\Encoder}} & 2$\times$ Conv2D 1$\times$1$\times$32 & (62$\times$62$)\times$32 \\
&  Conv2D 7$\times$7$\times$64 & (31$\times$31)$\times$64\\
&2$\times$ Conv2D 5$\times$5$\times$96 & (23$\times$23)$\times$96\\
&2$\times$ Conv2D 5$\times$5$\times$128  & (15$\times$15)$\times$128\\
&2$\times$ Conv2D 3$\times$3$\times$256  & (5$\times$5)$\times$256\\
&2$\times$ Conv2D 3$\times$3$\times$384  & (1$\times$1)$\times$384\\
& Up + Conv2D 1$\times$1$\times$384 & (5$\times$5)$\times$384 \\
& Conv2D 3$\times$3$\times$384 & (5$\times$5)$\times$384\\
& Up + Conv2D 1$\times$1$\times$384 & (15$\times$15)$\times$384 \\
& Conv2D 3$\times$3$\times$384 & (15$\times$15)$\times$384\\
& Up + Conv2D 1$\times$1$\times$384 & (23$\times$23)$\times$384 \\
& Conv2D 3$\times$3$\times$384 & (23$\times$23)$\times$384\\
& Up + Conv2D 1$\times$1$\times$384 & (31$\times$31)$\times$384 \\
& Conv2D 3$\times$3$\times$384 & (31$\times$31)$\times$384\\
& Up + Conv2D 1$\times$1$\times$384 & (62$\times$62)$\times$384 \\
& Conv2D 3$\times$3$\times$384 & (62$\times$62)$\times$384\\
& 2$\times$ Conv2D 3$\times$3$\times$384 & (62$\times$62)$\times$384\\
& *Correlation & (62$\times$62)$\times$1\\
& *Conv2D 3$\times$3$\times$128 & (62$\times$62)$\times$128\\
& Concatenate (1 + 128 + 128) & (62$\times$62)$\times$257\\
& Sample Pyramid & \{(62$\times$62), (31$\times$31)\}$\times$257\\
& Squeeze and Concatenate & (31$\times$31)$\times$514\\ \midrule

\multirow{7}{*}{\shortstack{Uncertainty\\Predictor}}  & 2$\times$ Conv2D 1$\times$1$\times$128  & (31$\times$31)$\times$128 \\ 
& 4$\times$ Conv2D 5$\times$5$\times$64  & (15$\times$15)$\times$64 \\ 
& 2$\times$ Conv2D 3$\times$3$\times$64  & (5$\times$5)$\times$64 \\ 
& 2$\times$ Conv2D 3$\times$3$\times$128  & (1$\times$1)$\times$128 \\ 
& Linear 2  & 2 \\ 
& Softmax  &  1\\ 
& Parabola Function  & 1 \\ \midrule

\multirow{9}{*}{\shortstack{LSTM\\Displacement\\Predictor}}& 2$\times$ Conv2D 3$\times$3$\times$64  & (15$\times$15)$\times$64  \\
& 2$\times$ Conv2D 3$\times$3$\times$128  & (7$\times$7)$\times$128 \\
& ConvLSTM 3$\times$3$\times$128 & (7$\times$7)$\times$128\\ 
& 3$\times$ Conv2D 3$\times$3$\times$256  & (1$\times$1)$\times$256 \\
& Concatenate Hidden Vector  &  512\\
& 2$\times$Linear 256  &  256\\
& Concatenate Hidden Vector  & 512 \\
& Gate Layer 256  & 256 \\
& Linear 2  &  2\\
\midrule

\end{tabular}
\caption{\textbf{Detail architecture of event module.} * Layers are processed simultaneously.}
\label{tab:event_model}
\end{table}

\noindent{\textbf{Image Module.}}
In our image module, the grayscale image is encoded to feature map $\mathbf{F}_{img}$ of $1/8$ image size with dim $D_{F_{img}} = 128$, and then correlated in $L_{img} = 4$ levels and extracted to patch with size $P_{img} = 7$. The point position (x, y) is embedded in 128 dim. The iterative prediction iterates 3 times. We also use 3 points to generate a parabola for uncertainty mapping, which maps $\{0, 0.5, 1\}$ to $\{0, 1, 10\}$.  Detailed network information is listed in Tab.~\ref{tab:image_model}.

\begin{table}[tb]
\centering
\scriptsize
\setlength\tabcolsep{3pt}
\begin{tabular}{clc}
\toprule
                  & Layer       & Data Size                     \\ \bottomrule

\multirow{9}{*}{\shortstack{Pyramid\\Encoder}}&  Conv2D 7$\times$7$\times$64 & (H/2$\times$W/2)$\times$64\\
&2$\times$ Conv2D 3$\times$3$\times$64 & (H/2$\times$W/2)$\times$64\\
&2$\times$ Conv2D 3$\times$3$\times$96  & (H/4$\times$H/4)$\times$96\\
 &2$\times$ Conv2D 3$\times$3$\times$128  & (H/8$\times$H/8)$\times$128\\
&2$\times$ Conv2D 3$\times$3$\times$128  & (H/16$\times$H/16)$\times$128\\
& *Interpolate and Concatenate & \multirow{2}{*}{(H/8$\times$H/8)$\times$416} \\
& \quad (64 + 96 + 128 + 128) & \\
& Conv2D 3$\times$3$\times$256 & (H/8$\times$W/8)$\times$256\\
& Conv2D 1$\times$1$\times$128 & (H/8$\times$W/8)$\times$128\\\midrule

\multirow{7}{*}{\shortstack{Correlation\\Pyramid}}& Input Feature Map & (H/8$\times$W/8)$\times$128\\
& Average Pool 2$\times$2 & (H/16$\times$W/16)$\times$128\\
& Average Pool 2$\times$2 & (H/32$\times$W/32)$\times$128\\
& Average Pool 2$\times$2 & (H/64$\times$W/64)$\times$128\\
& *Sample 7$\times$7 & 4$\times$(7$\times$7)$\times$128\\
& Correlate & 4$\times$(7$\times$7)$\times$1\\
& Concatenate & (7$\times$7)$\times$4\\ \midrule

\multirow{5}{*}{\shortstack{Prediction\\Head}}& Flatten, Embedding and Concatenate All Data & \multirow{4}{*}{582}  \\
&\quad(2$\times$feature vectors + correlation map +&\\
&\quad embedded position + position)&\\
&\quad(2$\times$128 + 7$\times$7$\times$4 + 128 + 2)&\\
& MLP, depth 12, hidden dim 512 & 4 \\ \midrule

\end{tabular}
\caption{\textbf{Detail architecture of image module.} Layer with "*`` have input from all previous 4 layers.}
\label{tab:image_model}
\end{table}

\section{More Experiment Results} \label{More Experiment Results}

\subsection{More Quantitative Comparisons} \label{More Quantitative Comparisons}

\begin{table*}[tb]
\centering
\footnotesize
\setlength\tabcolsep{10pt}
\begin{tabular}{c||l||ccc||ccc}
\toprule
\multirow{2}{*}{Data}  &\multirow{2}{*}{Methods} & \multicolumn{3}{c||}{EC-occ} & \multicolumn{3}{c}{EDS-occ} \\ 
& & $\delta_{avg}^{vis}\uparrow$ & $\delta_{avg}^{occ}\uparrow$ & $\delta_{avg}\uparrow$ & $\delta_{avg}^{vis}\uparrow$ & $\delta_{avg}^{occ}\uparrow$ & $\delta_{avg}\uparrow$ \\ \bottomrule
&HASTE\cite{haste}                              & 9.7           & 1.8           & 9.1           & 0.0           & 0.0           & 0.0           \\
&EKLT\cite{eklt}                           & 21.7          & 9.9           & 20.6          & 16.9          & 6.3           & 16.3          \\
Event&Deep-EV-Tracker\cite{deepev}              & 26.8          & 20.2          & 26.3          & 31.3          & 21.3          & 30.9          \\
&\textbf{Ours(Event)}                      & \underline{29.3}    & \underline{23.1}    & \underline{28.7}    & \underline{35.1}    & \underline{24.6}    & \underline{34.6}    \\
&\textbf{Ours(Event+Kalman)}               & \textbf{30.5} & \textbf{24.1} & \textbf{29.8} & \textbf{36.7} & \textbf{25.9} & \textbf{36.2} \\ \midrule

\multirow{4}{*}{\shortstack{Event\\+\\Image}}&FF-KDT\cite{ffkdt}   & 28.2        & 14.5         & 27.1    & 24.0          & 11.5          & 23.4            \\
&Deep-EV-Tracker\cite{deepev} + KLT\cite{klt} & 32.1          & 5.8           & 30.3          & 35.0          & 10.7          & 33.7          \\
&\textbf{Ours(Event+Image)}                & \underline{37.2}    & \underline{11.4}    & \underline{35.5}    & \underline{43.1}    & \underline{18.6}    & \underline{41.6}    \\
&\textbf{Ours(Event+Image+Kalman)}         & \textbf{44.5} & \textbf{28.5} & \textbf{43.4} & \textbf{52.0} & \textbf{26.8} & \textbf{50.6}\\ \midrule

\end{tabular}
\caption{\textbf{Performance evaluation on synthetic data with occlusion.}}
\label{tab:epe}
\end{table*}

\begin{table*}[tb]
\centering
\footnotesize
\setlength\tabcolsep{7.5pt}
\begin{tabular}{c||l||ccc||ccc}
\toprule
\multicolumn{8}{c}{Expect FA$\uparrow$} \\ \midrule
\multirow{2}{*}{Data}&\multirow{2}{*}{Methods} & \multicolumn{3}{c||}{EC} & \multicolumn{3}{c}{EDS} \\ 
& & EC-syn & EC-occ & Dcre.(\%)$\downarrow$ & EDS-syn & EDS-occ & Dcre.(\%)$\downarrow$\\ \bottomrule
&HASTE\cite{haste}                              & 0.379                & 0.341                & 10.0$\dagger$                 & 0.043                & 0.085                & -96.9$\dagger$                \\
&EKLT\cite{eklt}                           & 0.801                & 0.370                & 53.8                 & 0.353                & 0.324                & \textbf{8.3}         \\
Event&Deep-EV-Tracker\cite{deepev}              & 0.812                & 0.423                & 47.9                 & 0.454                & 0.289                & 36.2                 \\
&\textbf{Ours(Event)}                      & \textbf{0.830}       & \underline{0.522}          & \underline{37.2}           & \textbf{0.461}       & \underline{0.343}          & 25.5                 \\
&\textbf{Ours(Event+Kalman)}               & \underline{0.828}          & \textbf{0.527}       & \textbf{36.3}        & \underline{0.459}          & \textbf{0.349}       & \underline{23.9}           \\\midrule

\multirow{4}{*}{\shortstack{Event\\+\\Image}}&FF-KDT\cite{ffkdt} & 0.846*                & 0.401*                & 52.6*                 & 0.434*                & 0.212*                & 51.2*                 \\
&Deep-EV-Tracker\cite{deepev} + KLT\cite{klt} & 0.738                & 0.314                & 57.4                 & 0.504                & 0.274                & 45.6                 \\
&\textbf{Ours(Event+Image)}                & \underline{0.780}          & \underline{0.446}          & \underline{42.8}           & \underline{0.532}          & \underline{0.292}          & \underline{45.0}           \\
&\textbf{Ours(Event+Image+Kalman)}         & \textbf{0.845}       & \textbf{0.572}       & \textbf{32.4}        & \textbf{0.550}       & \textbf{0.356}       & \textbf{35.2}  \\ \midrule  
\end{tabular}
\caption{\textbf{Performance degradation on occluded data from non-occluded data.} $\dagger$ HASTE\cite{haste} exhibits such poor performance that its decreased value becomes meaningless. * FF-KDT\cite{ffkdt} can only produce estimates at image timestamps, whereas Deep-Ev-Tracker\cite{deepev} and our method generate significantly more predictions; therefore, it is excluded from our comparison.}
\label{tab:occ}
\end{table*}

\noindent{\textbf{Performance Experiment on Occlusion Data.}}
We evaluate accuracy through $\delta_{avg}$\cite{tapvid} on more methods.
The result shows our multi-modality method outperforms all other baselines on both $\delta_{avg}^{vis}$, $\delta_{avg}^{occ}$, and $\delta_{avg}$, and our event module with Kalman filter also outperforms all other event-based baselines, see Tab.~\ref{tab:epe}. 
The integration of a Kalman filter significantly enhances its performance in handling occluded points with minimal computational overhead, highlighting the Kalman filter's robust capability in effectively addressing occlusion.

\noindent{\textbf{Performance Degradation Experiment on Occlusion Data.}}
To assess the capability to handle occlusion, we conducted a series of experiments examining the impact of occlusion. We evaluated methods on synthetic and occluded versions of EC\cite{ec} and EDS\cite{eds} datasets outlined in Tab.~\ref{tab:occ}. The numerical decrease in performance from synthetic to occluded data provides insights into each method's robustness against occlusion.
The results indicate that, compared with Deep-EV-Tracker\cite{deepev}, our event module demonstrates greater robustness in handling occlusions due to the larger pyramid patch and enhanced encoder. Moreover, a significant improvement is observed when incorporating the Kalman filter and the uncertainty predictor.
The modules incorporating the Kalman filter exhibit less performance degradation across all data modalities. This finding further verifies the Kalman filter's strong capability in handling occlusions.
Notably, EKLT\cite{eklt} experiences the least performance degradation from EDS-syn to EDS-occ, which can be attributed to its already limited performance on EDS-syn.

\begin{table}[tb]
\centering
\scriptsize
\setlength\tabcolsep{4.1pt}
\renewcommand\arraystretch{1}
\begin{tabular}{c||l||ccc||ccc}
\toprule
\multirow{2}{*}{Data} & \multirow{2}{*}{Methods} & \multicolumn{3}{c||}{EC} & \multicolumn{3}{c}{EDS} \\ 
 & & FA$\uparrow$ & Exp FA$\uparrow$ & $N_{p}$$\uparrow$ & FA$\uparrow$ & Exp FA$\uparrow$ & $N_{p}$$\uparrow$ \\ \bottomrule

 &KLT\cite{klt}              & 0.734          & 0.729     &24     & 0.588          & 0.497     &75     \\
I& \textbf{Ours(I)}       & \underline{0.778}          & \underline{0.772}   &24       & \textbf{0.633} & \textbf{0.524}&75 \\
& \textbf{Ours(I w. K)}        & \textbf{0.784} & \textbf{0.778} & 24&\underline{0.619}          & \underline{0.511}  &75        \\ \midrule

\end{tabular}
\caption{\textbf{Performance evaluation on EC and EDS.}``I'' denotes image. $N_{p}$ denotes the number of predictions per second of data.}
\label{tab:eceds-image}
\end{table}
\noindent{\textbf{Performance Experiment on Image Module.}}
We also report the performance of the image-only model in Tab.~\ref{tab:eceds-image}, noting that it is designed to be lightweight. The performance drop observed when adding the Kalman filter on EDS is attributed to the domain gap between the training data and the EDS dataset, particularly due to motion blur, which is not yet modeled in the training data, our MultiTrack dataset. Nevertheless, the significant performance gain from integrating the event module, image module, and Kalman filter demonstrates that the Kalman filter effectively fuses the two modalities, leveraging the strengths of both.

\subsection{More Qualitative Comparison} \label{More Qualitative Comparison}
\begin{figure*}[tb]
  \centering
  \includegraphics[width=0.8\linewidth]{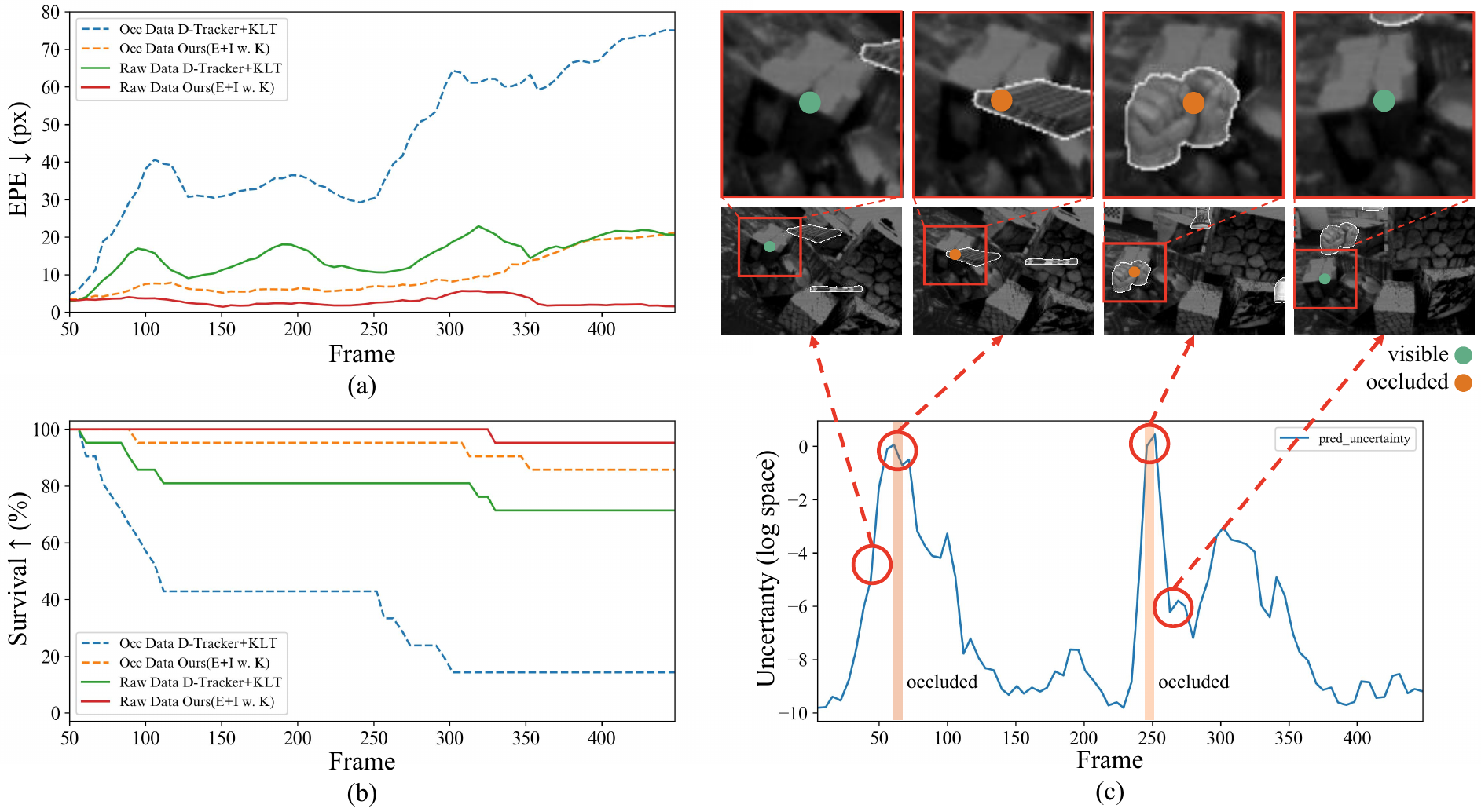}
  \caption{
  \textbf{Long-term stability experiment and uncertainty visualization.} 
  (a) Tracking quality (EPE). (b) Survival ratio. 
  (c) Uncertainty.
  }
  \label{fig:timeline}
\end{figure*}

\noindent{\textbf{Qualitative Comparison in Long-tern Stability.}}
Since our method is trained on sequences of up to 23 frames, we conduct a long-term stability experiment to evaluate its ability to generalize to longer sequences. The experiment is performed on Boxes Rotation from EC\cite{ec} and EC-occ, consisting of 81 image frames and 368 event frames. Metrics are computed only on frames with ground truth. Tracking performance is reported using end-point error (EPE) and survival ratio, where a track is considered non-surviving if the L2 distance exceeds 32 pixels.
From Fig.~\ref{fig:timeline} (a) and (b), our method demonstrates robust long-term stability and consistently outperforms Deep-Ev-Tracker\cite{deepev}+KLT\cite{klt} by a substantial margin on both raw and occluded data.

\begin{figure}[tb]
  \centering
  \includegraphics[width=1\linewidth]{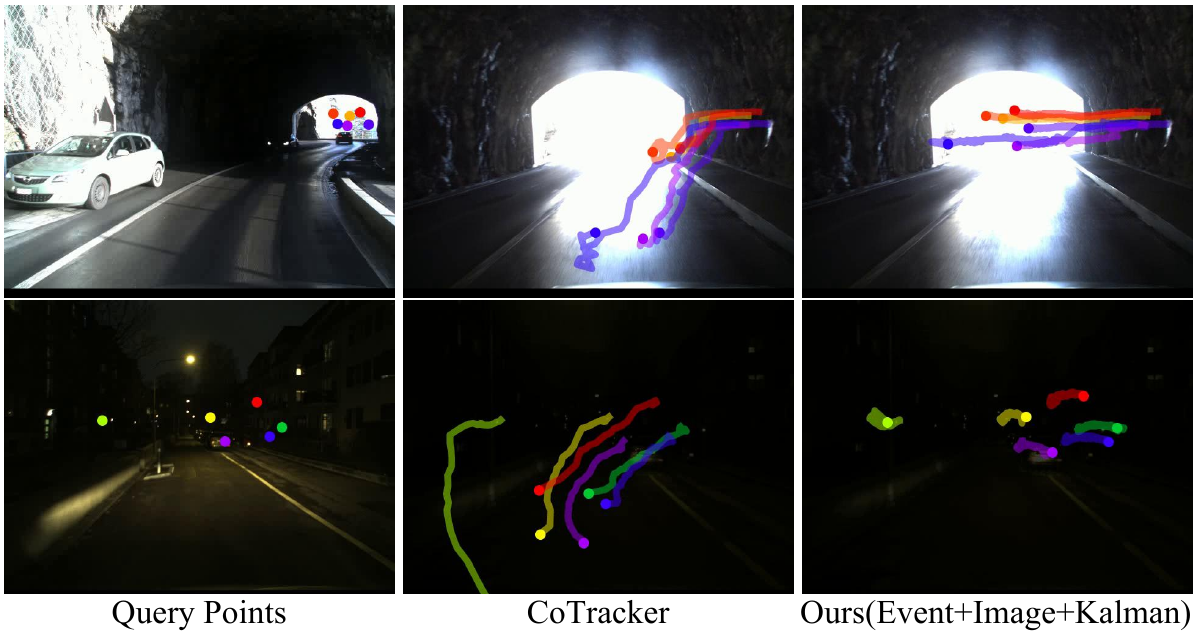}
  \caption{\textbf{Qualitative comparison with pure image-based method in extreme light condition.}}
  \label{fig:cotracker}
\end{figure}

\noindent{\textbf{Qualitative Comparison in Extreme Light Condition.}}
To demonstrate the superiority of event cameras, we present a qualitative comparison with the state-of-the-art image-based method, CoTracker\cite{cotracker}, in Fig~\ref{fig:cotracker}. The two scenes represent overexposure and underexposure scenarios, where the event camera remains unaffected. In contrast, CoTracker fails completely, while our method continues to operate, highlighting the essential role of event cameras in extreme conditions alongside image data. Both scenes are from DSEC\cite{dsec}, with the first using raw data and the second manually adjusted to simulate extreme underexposure.

\begin{figure*}[tb]
  \centering
  \includegraphics[width=1\linewidth]{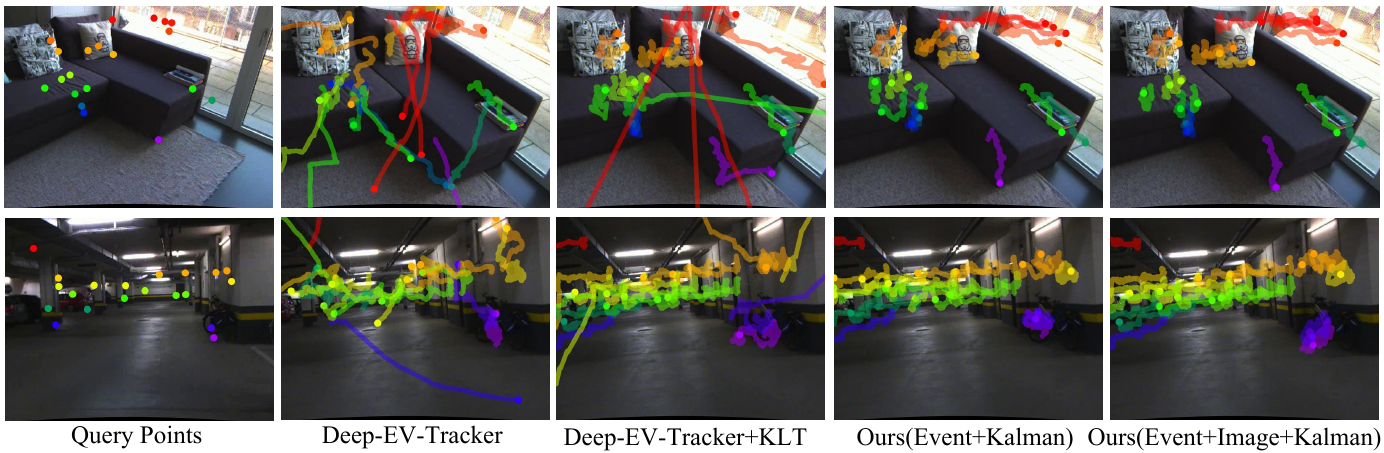}
  \caption{\textbf{Qualitative comparison.}}
  \label{fig:result_supp}
\end{figure*}

\noindent{\textbf{More Qualitative Comparison.}}
In this section, we present more qualitative results (Fig.~\ref{fig:result_supp}). Both scenes are from EDS\cite{eds}. The first scene features a sofa with low texture, which challenges the event tracker, while the second scene suffers from insufficient lighting, posing difficulties for the image tracker. 
The results demonstrate that our method achieves superior performance, effectively addressing the limitations of both trackers through the integration of the Kalman filter.

\subsection{Uncertainty Visualization}  \label{Uncertainty Visualization}
To evaluate the effectiveness of uncertainty training, we visualize the uncertainty during tracking in Fig.~\ref{fig:timeline} (c). The experiment is conducted on Boxes Rotation from EC-occ, where we select a track that includes both visible and occluded states. The results show increased uncertainty during challenging scenarios (e.g., occlusion), demonstrating the effectiveness of our uncertainty modeling.

\subsection{More Runtime Analysis} \label{More Runtime Analysis}

\begin{figure}[tb]
  \centering
  \includegraphics[width=1\linewidth]{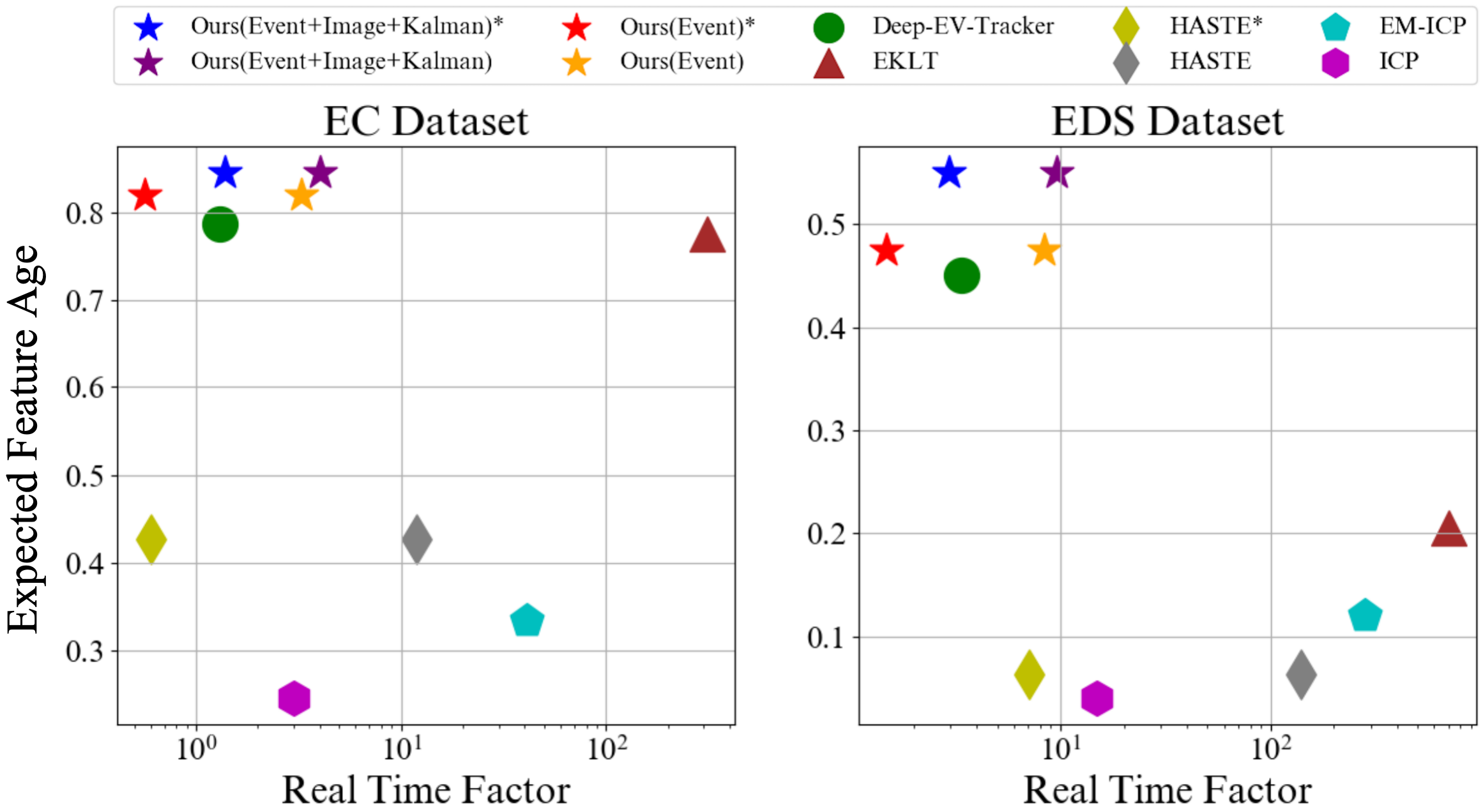}
  \caption{\textbf{Runtime performance in terms of expected feature age with the real-time factor.} Methods that exhibit superior performance and computational efficiency tend to position closer to the upper left corner. The statics of EKLT\cite{eklt}, HASTE\cite{haste}, EM-ICP\cite{emicp}, ICP\cite{icp} are from\cite{deepev}. Deep-Ev-Tracker(E2VID)\cite{deepev2} has the same architecture as Deep-Ev-Tracker, so they should have the same runtime performance. Methods with ``*'' assume perfect parallelization for processing all feature tracks. Our method without ``*'' is implemented with unoptimized parallelization. The real-time factor is simply the computation time divided by input data interval.\cite{deepev}.
  }
  \label{fig:runtime}
\end{figure}
\noindent{\textbf{Runtime and Accuracy Performance}}
Additional experiments related to runtime are presented in Fig.~\ref{fig:runtime}. The statistics of other methods are from\cite{deepev}, and our experiment is conducted on the same type of GPU. The real-time factor is simply the computation time divided by input data interval.\cite{deepev}. 
Our module does not have dependencies between event patches, supporting fully parallel calculation. The perfect parallelization results were obtained by tracking one feature with one thread across multiple event frames. On the contrary, the Deep-EV-Tracker requires information exchange and motion consistency across tracks, thus not supporting perfect parallel calculation. 

It can be observed from Fig.~\ref{fig:runtime} that: 

(1) Our event module achieves the best runtime performance and accuracy on both datasets. 

(2) With the assistance of the image module and the Kalman filter, our method achieves almost the same runtime performance as Deep-EV-Tracker\cite{deepev} while getting much better tracking performance.

(3) In practice, our methods without perfect parallelization still exhibit a superior runtime compared to most existing methods. Further optimization of parallel computation could enhance performance.

\begin{table}[tb]
\centering
\scriptsize
\setlength\tabcolsep{7.6pt}
\renewcommand\arraystretch{1}
\begin{tabular}{l||l||cc||c}
\toprule
\multicolumn{5}{c}{Runtime(ms)$\downarrow$} \\ \midrule
Experiment & Method & EC & EDS & Paras \\ \bottomrule
Feature Pyramid        & \underline{On}                                 & 7.95 & 8.70 & 33.1M \\
(Ours Event) & Off     & 5.88 & 6.91 & 29.3M \\\midrule
LSTM Module   & \underline{Feature+Displace}    & 7.95 & 8.70 & 33.1M \\
(Ours Event)  & Feature  & 6.55 & 7.79 & 32.7M    \\
                       & Displacement   & 6.54 & 8.16 & 31.9M    \\
                       & Off  & 6.04 & 6.75 & 31.6M \\\midrule
Correlation Vector     & \underline{Feature Map Center}                  & 7.95 & 8.70 & 33.1M \\
(Ours Event) & U-Net Bottleneck & 8.17 & 9.04 & 35.7M    \\\midrule

\end{tabular}
\caption{\textbf{Ablation settings runtime.}}

\label{tab:ablation_runtime}
\end{table}
\noindent{\textbf{Runtime of Ablation Settings}}
To analyze the runtime of each component under different ablation settings, we report the runtime results in Table~\ref{tab:ablation_runtime}. Disabling the feature pyramid reduces runtime by 2ms but leads to significant performance degradation. Removing each LSTM module saves approximately 1ms, at the cost of substantial accuracy loss. Using the feature map center as the correlation vector improves both runtime and performance.

\subsection{Event Module Reference Frame Experiment} \label{Event Module Reference Frame Experiment}

\begin{figure}[tb]
  \centering
  \includegraphics[width=1\linewidth]{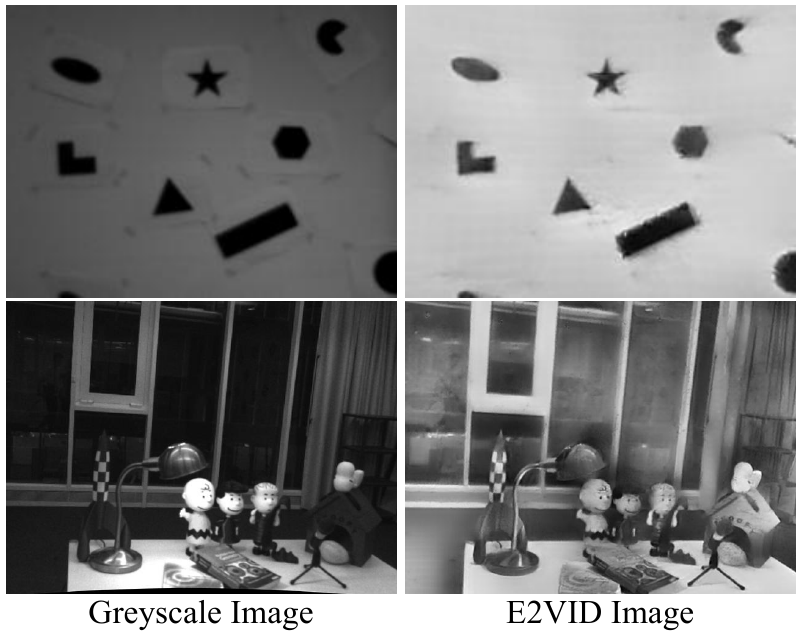}
  \caption{\textbf{Comparison of greyscale image and E2VID\cite{e2vid, e2vid2} image.}}
  \label{fig:e2vid}
\end{figure}

Our event module uses a greyscale image as the reference frame to ensure that both the event and image modules track the same feature points in the same reference frame, eliminating reference frame bias. 
Although tracking with the event module requires additional spatial alignment between the two cameras, our primary contribution lies in integrating both event and image data for tracking. Thus, spatial alignment should be considered a preliminary step rather than an additional burden.
Moreover, using different modality reference frames requires synchronization at a specific reference timestamp, which is incompatible with our asynchronous tracking framework.
We must note that previous methods, such as EKLT\cite{eklt} and Deep-Ev-Tracker\cite{deepev}, also use grayscale images as reference frames, as image data provides richer information. In contrast, event data is limited in low-texture or stationary scenes.

However, our event module can also track using pure event data. By employing event-to-frame reconstruction with E2VID\cite{e2vid, e2vid2}, as shown in Fig.~\ref{fig:e2vid}, we can replace the greyscale reference image with the reconstructed image\cite{deepev2}.

\section{More Details about MultiTrack, EC-occ and EDS-occ} \label{More Details about MultiTrack, EC-occ and EDS-occ}

\subsection{Displacement Distribution of MultiFlow and MultiTrack} \label{Displacement Distribution of MultiFlow and MultiTrack}

\begin{figure}[tb]
  \centering
  \includegraphics[width=1\linewidth]{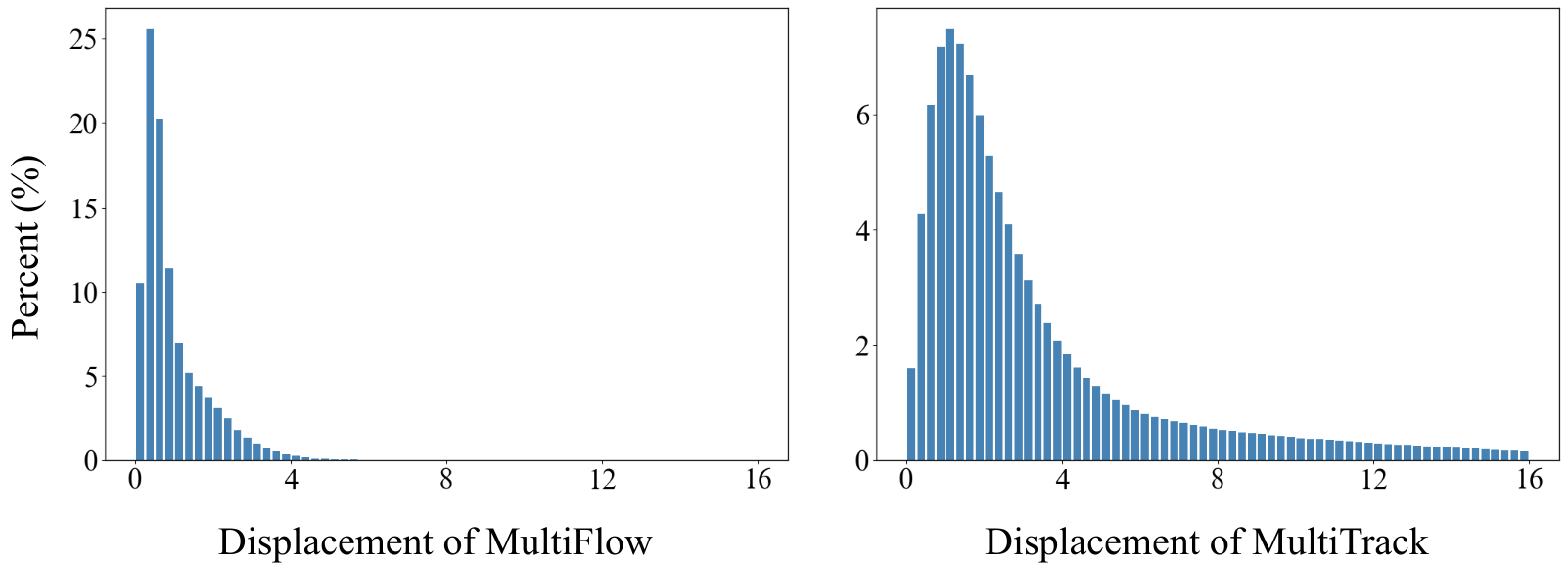}
  \caption{\textbf{Ground truth displacement distribution of MultiFlow\cite{multiflow} and MultiTrack.} }
  \label{fig:distribution}
\end{figure}

It is established that trajectory error tends to increase with greater displacement in the trajectory\cite{multiflow}. Based on this, we analyzed the ground truth displacement statistics between adjacent event frames to illustrate the level of difficulty, see Fig.~\ref{fig:distribution}.
The results indicate that MultiFlow\cite{multiflow} has most displacements below 4 pixels, whereas MultiTrack exhibits significantly larger displacements and point occlusions. These factors increase complexity, suggesting that MultiTrack presents a greater challenge than MultiFlow.

Our investigation suggests that datasets with appropriate difficulty levels are essential for unlocking the potential of learning-based modules. Therefore, MultiTrack plays a crucial role in addressing the shortage of event-based feature tracking datasets by offering adjustable synthetic parameters to control difficulty levels.

\subsection{Data Quality of MultiFlow and MultiTrack}\label{Data Quality of MultiFlow and MultiTrack}

\begin{table}[tb]
\centering
\scriptsize
\setlength\tabcolsep{11.3pt}
\renewcommand\arraystretch{1}
\begin{tabular}{l||cc||cc}
\toprule
\multicolumn{5}{c}{Expect FA$\uparrow$} \\ \midrule
 Dataset & EC & EC-occ & EDS & EDS-occ \\ \bottomrule
 \underline{MultiTrack}                 & \textbf{0.800} & \textbf{0.510} & \textbf{0.412} & \textbf{0.283}  \\
MultiFlow\cite{multiflow}                         & 0.777 & 0.453 & 0.405 & 0.269  \\\midrule
\end{tabular}
\caption{\textbf{Training dataset comparisom.}}

\label{tab:dataset}
\end{table}
In this experiment, we train our event module and Deep-EV-Tracker\cite{deepev} respectively on the MultiTrack and MultiFlow\cite{multiflow} datasets and evaluate the average performance of the two methods on the two datasets. Although we view MultiTrack as a supplement to MultiFlow rather than a complete replacement, the results in Tab.~\ref{tab:dataset} show that our MultiTrack dataset still outperforms the MultiFlow dataset on average metrics. This suggests that MultiTrack has better data quality and can unleash the full potential of modules.

\subsection{Synthetic Data Quality of EC-occ and EDS-occ}\label{Synthetic Data Quality of EC-occ and EDS-occ}

\begin{table}[tb]
\centering
\scriptsize
\setlength\tabcolsep{9pt}
\renewcommand\arraystretch{1}
\begin{tabular}{l||cc||cc}
\toprule
\multicolumn{5}{c}{Expect FA$\uparrow$} \\ \midrule

Methods & EC & EC-syn & EDS & EDS-syn \\ \bottomrule

Deep-EV-Tracker\cite{deepev} & 0.787          & 0.812          & 0.451          & 0.454          \\
\textbf{Ours(Event)}   & \textbf{0.819} & \textbf{0.830} & \textbf{0.474} & \textbf{0.461}  \\ \midrule
\end{tabular}
\caption{\textbf{Performance evaluation on real event data and synthetic event data.}}
\label{tab:syn_real}
\end{table}
Some hold that synthetic events exhibit significant differences from events captured by real event cameras. 
So we experiment by also generating events from interpolated but not occluded images, EC-syn and EDS-syn, to measure the real events and our synthetic events. Tab.~\ref{tab:syn_real} shows the slight difference in evaluation metric between real and synthetic data, which proves that the synthetic data could be credible in evaluation.

\subsection{Visualization Examples of MultiTrack, EC-occ and EDS-occ} \label{Visualization Examples of MultiTrack, EC-occ and EDS-occ}
\begin{figure*}[tb]
  \centering
  \includegraphics[width=1\linewidth]{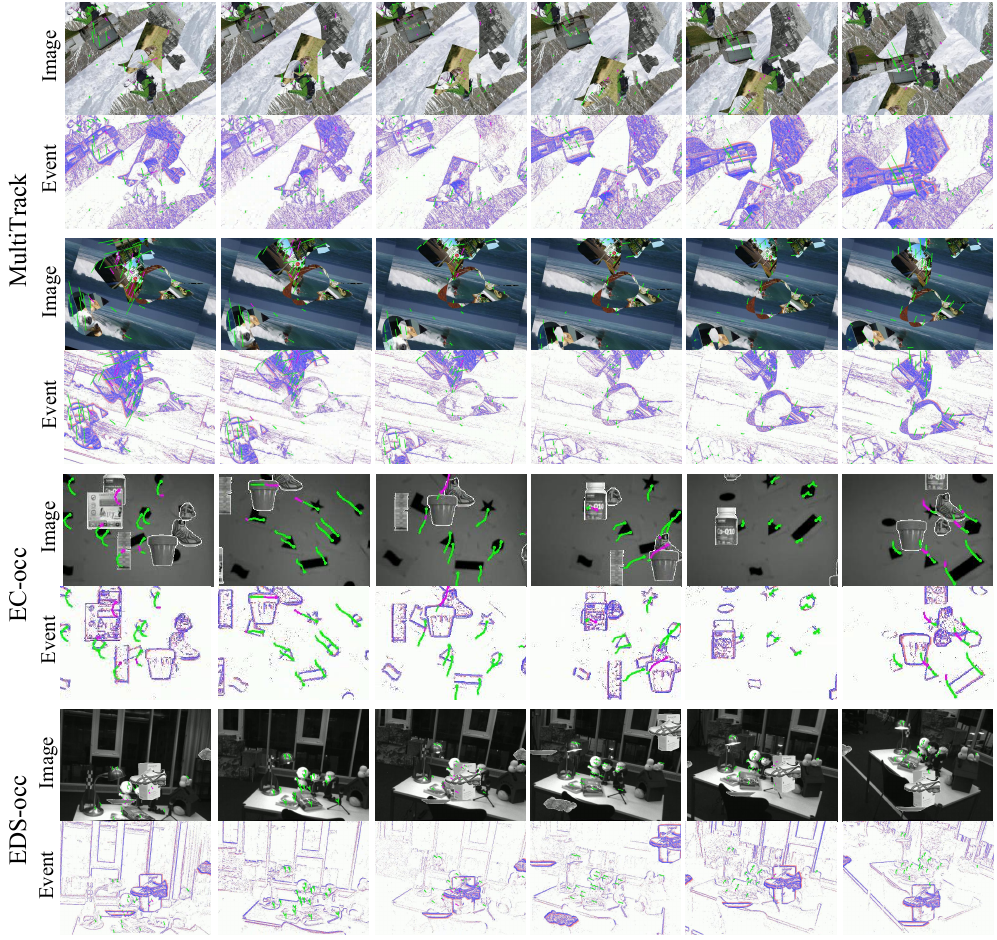}
  \caption{\textbf{Visualization examples of MultiTrack, EC-occ and EDS-occ.}}
  \label{fig:dataset}
\end{figure*}

We provide some visualization examples in Fig.~\ref{fig:dataset}. The ground truth point tracks are depicted as trajectories, with green indicating that the points are visible and purple indicating that the points are occluded at that moment.

\subsection{Dataset License} \label{Dataset License}
In this section, we present the licensing information for the datasets we used, as well as for our generated datasets, including EC-occ, EDS-occ, and the MultiTrack dataset.
The data we use are all from existing datasets, among them Google Scanned Objects\cite{googlescanned}is under Creative Commons Attribution 4.0 License\cite{cc40} which are free to share and adapt, while Flickr30k\cite{flickr30k} and COCO2014\cite{coco} are under Flickr Terms of Use\cite{flickrterms} which is available for researchers and educators who wish to use the dataset for non-commercial research or educational purposes. According to these, the EC-occ and EDS-occ are under Creative Commons Attribution 4.0 License, free to use by anyone, while using the open-source version of MultiTrack must abide by the Flickr Terms of Use. However, our dataset generator could replace the Flickr30k or COCO2014 with other image datasets with a Creative Commons Attribution 4.0 License, producing data friendly to commercial use.

\newpage

\cleardoublepage
{
    \small
    \bibliographystyle{ieeenat_fullname}
    \bibliography{main}
}

\end{document}